%% file: main.tex
\documentclass{article}

\usepackage{microtype}
\usepackage{graphicx}
\usepackage{subfigure}
\usepackage{booktabs} 

\usepackage{hyperref}

\usepackage[accepted]{icml2023}

\usepackage{amsmath}
\usepackage{amssymb}
\usepackage{mathtools}
\usepackage{amsthm}
\usepackage{enumitem}

\usepackage[capitalize,noabbrev]{cleveref}

\newcommand{\bfm}{\boldsymbol{m}}
\newcommand{\bfp}{\boldsymbol{p}}
\newcommand{\bfs}{\boldsymbol{s}}
\newcommand{\bfxi}{\boldsymbol{\xi}}
\newcommand{\bfsigma}{\boldsymbol{\sigma}}
\newcommand{\bfA}{\boldsymbol{A}}
\newcommand{\calF}{\mathcal{F}}
\newcommand{\calM}{\mathcal{M}}
\newcommand{\calA}{\mathcal{A}}
\newcommand{\calB}{\mathcal{B}}
\newcommand{\calC}{\mathcal{C}}
\newcommand{\calP}{\mathcal{P}}
\newcommand{\calD}{\mathcal{D}}
\newcommand{\calH}{\mathcal{H}}
\newcommand{\calG}{\mathcal{G}}
\newcommand{\calS}{\mathcal{S}}
\newcommand{\calI}{\mathcal{I}}
\newcommand{\calJ}{\mathcal{J}}
\newcommand{\calQ}{\mathcal{Q}}
\newcommand{\calE}{\mathcal{E}}
\newcommand{\calU}{\mathcal{U}}
\newcommand{\single}{\text{single}}
\newcommand{\policy}{\text{policy}}

\newcommand{\bbN}{\mathbb{N}}
\newcommand{\bbE}{\mathbb{E}}
\newcommand{\bbP}{\mathbb{P}}
\newcommand{\bbI}{\mathbb{I}}
\newcommand{\bbR}{\mathbb{R}}
\newcommand{\ee}{\text{ee}}

\DeclareMathOperator{\kl}{kl}
\DeclareMathOperator{\mo}{mod}
\DeclareMathOperator{\regret}{Reg}
\DeclareMathOperator{\reward}{Rew}
\DeclareMathOperator{\noneq}{NonEqu}
\DeclareMathOperator{\ber}{ber}
\DeclareMathOperator{\poa}{PoA}
\DeclareMathOperator{\round}{round}

\makeatletter
\def\maketag@@@#1{\hbox{\m@th\normalfont\normalsize#1}}
\makeatother

\usepackage[capitalize,noabbrev]{cleveref}

\theoremstyle{plain}
\newtheorem{theorem}{Theorem}[section]
\newtheorem{proposition}[theorem]{Proposition}
\newtheorem{lemma}[theorem]{Lemma}
\newtheorem{corollary}[theorem]{Corollary}
\theoremstyle{definition}
\newtheorem{definition}{Definition}[section]
\newtheorem{assumption}{Assumption}[section]

\newtheorem{example}{Example}[section]
\theoremstyle{remark}
\newtheorem{remark}{Remark}[section]

\usepackage[textsize=tiny]{todonotes}

\usepackage{macros}

\newenvironment{itm}
{\begin{itemize}[noitemsep,topsep=0pt,parsep=0pt,partopsep=0pt]}
{\end{itemize}}

\icmltitlerunning{Competing for Shareable Arms in Multi-Player Multi-Armed Bandits}

\begin{document}

\twocolumn[
\icmltitle{Competing for Shareable Arms in Multi-Player Multi-Armed Bandits}



\icmlsetsymbol{equal}{*}

\begin{icmlauthorlist}
\icmlauthor{Renzhe Xu}{equal,thucs}
\icmlauthor{Haotian Wang}{equal,thucs}
\icmlauthor{Xingxuan Zhang}{thucs}
\icmlauthor{Bo Li}{thusem}
\icmlauthor{Peng Cui}{thucs}
\end{icmlauthorlist}

\icmlaffiliation{thucs}{Department of Computer Science and Technology, Tsinghua University, Beijing, China}
\icmlaffiliation{thusem}{School of Economics and Management, Tsinghua University, Beijing, China. Emails: xrz199721@gmail.com, accwht@hotmail.com, xingxuanzhang@hotmail.com, libo@sem.tsinghua.edu.cn, cuip@tsinghua.edu.cn}

\icmlcorrespondingauthor{Peng Cui}{cuip@tsinghua.edu.cn}

\icmlkeywords{Multi-Armed Multi-Player Bandit, Strategic}

\vskip 0.3in
]



\printAffiliationsAndNotice{\icmlEqualContribution} 

\input{paragraphs/abstract.tex}
\input{paragraphs/introduction.tex}
\input{paragraphs/preliminary.tex}
\input{paragraphs/method.tex}
\input{paragraphs/experiments.tex}

\input{paragraphs/related_works.tex}
\input{paragraphs/conclusion.tex}

\clearpage

\bibliography{references}
\bibliographystyle{icml2023}

\clearpage
\appendix
\onecolumn

\input{paragraphs/appendix-additional}
\input{paragraphs/appendix-experiment}
\input{paragraphs/appendix-proof.tex}
\input{paragraphs/appendix-relaxed-setting.tex}
\input{paragraphs/appendix-auxiliary.tex}

\end{document}

%% file: paragraphs/abstract.tex
\begin{abstract}
    Competitions for shareable and limited  resources have long been studied with strategic agents. In reality, agents often have to learn and maximize the rewards of the resources at the same time. To design an individualized competing policy, we model the competition between agents in a novel multi-player multi-armed bandit (MPMAB) setting where players are selfish and aim to maximize their own rewards. In addition, when several players pull the same arm, we assume that these players averagely share the arms' rewards by expectation. Under this setting, we first analyze the Nash equilibrium when arms' rewards are known. Subsequently, we propose a novel \textbf{S}elfish \textbf{M}PMAB with \textbf{A}veraging \textbf{A}llocation (SMAA) approach based on the equilibrium. We theoretically demonstrate that SMAA could achieve a good regret guarantee for each player when all players follow the algorithm. Additionally, we establish that no single selfish player can significantly increase their rewards through deviation, nor can they detrimentally affect other players' rewards without incurring substantial losses for themselves.  We finally validate the effectiveness of the method in extensive synthetic experiments.
\end{abstract}

%% file: paragraphs/introduction.tex
\section{Introduction}
Agents compete with each other to maximize their own rewards in various applications with shareable and limited resources, ranging from cognitive radio networks \citep{mitola1999cognitive} to algorithm-curated platforms \citep{hron2022modeling}. For example, secondary users compete for channels to transmit data in cognitive radio networks \citep{mitola1999cognitive,haykin2005cognitive,meshkati2007energy}. Regarding algorithm-curated platforms (such as YouTube and TikTok), content providers create content with different topics to compete for exposure \citep{ben2020content,hron2022modeling,jagadeesan2022supply,yao2023bad,zhu2023online}. In these applications, resources are often shareable, and each resource has a limited total reward, such as the quality of each channel in cognitive radio networks and the demand for each content topic in algorithm-curated platforms. As a result, when more agents choose to acquire the same resource, the reward earned by each of them will decrease. Meanwhile, agents usually do not have prior knowledge about the total reward of each resource in reality~\citep{jouini2009multi,jouini2010upper,li2010contextual}. Therefore, these agents need to design policies to maximize their rewards faced with unknown resource rewards and competition from other agents.

It is a common practice to adopt the multi-player multi-armed bandit (MPMAB) framework \citep{anantharam1987asymptotically,liu2010distributed,anandkumar2011distributed,boursier2022survey} to model the behaviors of multiple agents with unknown rewards of different resources.
In this framework, players pull arms simultaneously at various rounds. Since several players may choose the same arm, different collision models \citep{bande2019multi,liu2020competing,shi2021multi,boyarski2021distributed,wang2022multi,wang2022multiple} are proposed to allocate the arm's reward to collided players. However, these works mainly focus on the cooperative player setting where players are non-strategic and cooperate to maximize the total reward of all players. This does not agree with the real scenarios since players are always selfish and target to maximize their own reward. Several works \citep{boursier2020selfish,liu2020competing,liu2021bandit,jagadeesan2021learning} consider the selfish behaviors of players while they assume that the resource is not shareable, as at most one player can get the reward when a collision occurs.

In this paper, we propose a novel MPMAB setting to characterize the strategic behaviors of agents in applications with limited and shareable resources.
We assume that the players are selfish, and they could strategically alter their policies to achieve better rewards.
In addition, when several players pull the same arm, we suppose that these players averagely share the arms' rewards by expectation.
We further assume that each player could observe his own reward and the reward of the chosen arm, similar to the statistic sensing setting in traditional MPMAB literature \citep{boursier2022survey}. Our target is to design an algorithm for each player that maximizes his own reward and is robust to the strategic behaviors of any other player. Specifically, the algorithm should achieve a good regret guarantee when all players follow the algorithm. Furthermore, no single selfish player can substantially increase their rewards through deviation, nor can they negatively impact other players' rewards without incurring significant personal losses.

To achieve this target, we first analyze the Nash equilibrium of players' behaviors when the expected reward of each arm is known. Based on the equilibrium, we propose a novel \textbf{S}elfish \textbf{M}PMAB with \textbf{A}veraging \textbf{A}llocation (SMAA) approach for each player that can explore the rewards of arms, maximize his own reward, and converge to equilibrium at the same time. We theoretically analyze our algorithm and demonstrate that after $T$ rounds, (1) when all players follow SMAA, the regret for each player is $O(\log T)$. We further show that it matches the lower bound when players only use the information of chosen arms' rewards. (2) When all players follow SMAA, the algorithm could converge to equilibrium, and the number of non-equilibrium rounds is $O(\log T)$. (3) A selfish player that deviates from the algorithm will bring at most $O(\log T)$ increase in his own reward. (4) If a selfish player wants to bring loss $u$ to another player's reward, he will also suffer from a loss of at least $\beta u - O(\log T)$, where $\beta$ is a constant. We finally validate the effectiveness of our method through extensive synthetic experiments.

To conclude, our contributions are listed as follows.

\begin{itm}
    \item We propose a novel MPMAB setting with an averaging allocation model to characterize the selfish behaviors of players in applications with limited and shareable resources.
    \item We analyze the Nash equilibrium of the problem and further propose a novel Selfish MPMAB with Averaging Allocation (SMAA) approach for players under the proposed setting.
    \item We theoretically demonstrate that SMAA could achieve a good regret guarantee for each player when all players follow the algorithm and it is robust to a single player's strategic deviation.
\end{itm}


%% file: paragraphs/preliminary.tex
\section{Preliminaries} \label{sect:preliminaries}
\subsection{Basic Setting}
Suppose there are $K$ arms with indices $[K] = \{1, 2, \dots, K\}$, $N$ players (he) and $T$ rounds.
At round $t \in [T]$, the reward of arm $k$ is given by $X_k(t)$ drawn \textit{i.i.d.} according to a distribution $\xi_k \in \Xi$ with expectation $\mu_k > 0$ supported on $[0,1]$. Here $\Xi$ represents a set of possible distributions supported on $[0, 1]$  with positive expectations.

The problem is decentralized. At each round $t \in [T]$, $N$ players choose arms simultaneously based only on their historical observations. We use $\pi_j(t)$ to denote the arm chosen by player $j$ and $M_k(t) = \sum_{j=1}^N \bbI[\pi_j(t) = k]$ to represent the number of players that pull arm $k$ at round $t$.

We assume all players are homogeneous, leading to an averaging allocation mechanism when collisions occur.
Formally, at round $t$, each player $j$ is endowed with a weight $w_j(t)$ sampled \textit{i.i.d.} from a distribution $\Gamma$ supported on $[0, 1]$. Then each arm's reward is allocated proportionally to the players' weights. Let random variable $R_j(t)$ denote the reward earned by player $j$ at round $t$, and it is given by the following equation:
\begin{equation}
    \small
    R_j(t) = X_{\pi_j(t)}(t) \cdot \frac{w_j(t)}{\sum_{j'=1}^N \bbI[\pi_{j'}(t) = \pi_j(t)]w_{j'}(t)}.
\end{equation}
As a result, since all players' weights are sampled from the same distribution $\Gamma$, we have that for all $k \in [K]$ such that $M_k(t) > 0$ and $j \in [N]$ such that $\pi_j(t) = k$,
\begin{equation}
    \small
    \bbE\left[R_j(t) | X_k(t), M_k(t)\right] = X_k(t) / M_k(t).
\end{equation}
This indicates that players averagely share the arms' rewards by expectation.

We consider the setting similar to the statistic sensing setting in standard multi-player multi-armed bandit problems \citep{boursier2022survey}. Formally, at each round, player $j$ could observe his own reward $R_j(t)$ and the total reward of the selected arm $\pi_j(t)$ (\textit{i.e.}, $X_{\pi_j(t)}(t)$).

\begin{example}
    We demonstrate one example of our proposed problem setting. In an algorithm-curated platform \citep{hron2022modeling}, such as YouTube and TikTok, content providers create content with different topics to compete for exposure. Here content providers are players and different content topics are arms. The reward of each arm corresponds to the total exposure to the topic. As the demand for each topic is usually stable \citep{ben2020content,hron2022modeling}, the expected amount of exposure for each topic is unchanged. Therefore, assuming players are homogeneous \citep{hron2022modeling,jagadeesan2022supply}, when several content providers select the same topic, the exposure is shared averagely among these providers by expectation.
\end{example}

\paragraph{Other mathematical notations}
We use $\floor{x}$ to denote the largest integer that is not greater than $x$ and $\ceil{x}$ to denote the smallest integer that is not smaller than $x$ for any real number $x$. For any $(p, q) \in [0, 1]^2$, let $\kl(p, q)$ denote the Bernoulli Kullback-Leibler divergence  defined as $\kl(p, q) = p \log (p/q) + (1-p) \log ((1 - p)/(1 - q))$ \citep{garivier2011kl}. In addition, with convention, we have that $0 \log 0 = 0 \log 0 / 0 = 0$ and $x \log x / 0 = \infty$ if $x > 0$.

\paragraph{Background on Nash equilibrium}
We adopt the definition of $\epsilon$-Nash equilibrium from the game theory \citep{nisan2007algorithmic}. For a general game problem, let $\calS$ be the set of agents' strategies, and each agent $j$ follows a strategy
$s_j \in \calS$. We use $\bfs = \{s_j\}_{j=1}^N$ to denote the strategy profile of all players and $(s', \bfs_{-j})$ to denote the profile where a single player $j$ deviates from the original strategy $s_j$ to a new strategy $s' \in \calS$. In addition, we use $\calU_j(\bfs)$ to denote the utility of player $j$ when the strategy profile is $\bfs$. With these notations, $\epsilon$-Nash equilibrium is defined as follows.

\begin{definition}[$\epsilon$-Nash equilibrium]
    For a game specified by $(\calS, \{\calU_j\}_{j=1}^N)$, a strategy profile $\bfs \in \calS^N$ is an $\epsilon$-Nash equilibrium if for all $s' \in \calS$ and $j \in [N]$,
    \begin{equation}
        \small
        \bbE\left[\calU_j(s', \bfs_{-j})\right] \le \bbE[\calU_j(\bfs)] + \epsilon.
    \end{equation}
\end{definition}

Moreover, if a strategy profile $\bfs$ is a $0$-Nash equilibrium, we say that the strategy profile $\bfs$ is a Nash equilibrium.

\subsection{Evaluation metrics} \label{sect:metrics}
In this paper, we target an algorithm for each player that maximizes his own reward and is robust to the strategic behaviors of any other player.
We first introduce two metrics to evaluate a policy's performance when all players follow the algorithm.

\paragraph{Players' regret}
Since we focus on the selfish player setting, we evaluate the regret of each player compared with his best choice in each round. At round $t$, if player $j$ pulls arm $k \in [K] \backslash \{\pi_j(t)\}$, the expected reward is $\mu_k / (M_k(t) + 1)$. If player $j$ pulls the originally chosen arm $\pi_j(t)$, his expected reward is $\mu_{\pi_j(t)} / M_{\pi_j(t)}(t)$. Combining these two cases, we can get the following equation for player $j$'s regret.
\begin{equation} \label{eq:regret}
    \small
    \regret_j(T) = \sum_{t=1}^T\left(\max_{k \in [K]}\frac{\mu_{k}}{M_{k}(t) + \bbI[\pi_j(t) \ne k]} - R_j(t)\right).
\end{equation}
Notably, each agent's regret depends on others agents' policies since $M_k(t)$ is the number of players on arm $k$.

\paragraph{Number of non-equilibrium rounds}
When players know the arms' expected rewards, their chosen arms will follow the Nash equilibrium demonstrated in \sectionref{sect:PNE}. As a result, we use the number of non-equilibrium rounds to evaluate whether an algorithm could effectively converge. Formally, the metric can be calculated by the following equation.
\begin{equation} \label{eq:non-equ}
    \small
    \noneq(T) = \sum_{t=1}^T \bbI\left[\exists k \in [K], M_k(t) \ne m^*_k\right],
\end{equation}
where $m^*_k$ is provided in \theoremref{thrm:PNE} and it represents the number of players that pull arm $k$ in the equilibrium.

We further provide two perspectives to analyze the policy's stability when there exists a single strategic player. We use $\calS_{\policy}$ to denote the set of all policies for players. For any $\bfs \in \calS_{\policy}^N$, the utility of player $j$ is the total reward after $T$ rounds, \textit{i.e.}, $\calU^{\policy}_j(\bfs) = \reward_j(T; \bfs) = \bbE\left[\sum_{t=1}^TR_j(t;\bfs)\right]$. Here we slightly abuse the notation and use $R_j(t;\bfs)$ to represent the reward earned by player $j$ at round $t$ when the policy profile is $\bfs$.

\paragraph{$\epsilon$-Nash equilibrium \textit{w.r.t.} policies} We expect a policy profile $\bfs$ is a $\epsilon$-Nash equilibrium \textit{w.r.t.} the game specified by $(\calS_{\policy}, \{\calU^{\policy}_j\}_{j=1}^N)$ with a small $\epsilon$ such that any single selfish player can not significantly improve his own reward by deviation.

\begin{remark}
    We note that the equilibrium mentioned here is not the equilibrium introduced when calculating the number of non-equilibrium rounds. Here the strategy set $\calS_{\policy}$ represents all policies and the utility function is the total reward after $T$ rounds. However, in the equilibrium mentioned in \equationref{eq:non-equ} (Details are provided in \sectionref{sect:PNE}), the strategy set is the set of all arms and the utility function is the expected reward at a single round.
\end{remark}

\paragraph{$(\beta, \epsilon)$-stable}
Following \citep{boursier2020selfish}, we further analyze how a single selfish player could affect other players' reward by the following definition.
\begin{definition}[$(\beta, \epsilon)$-stable \citep{boursier2020selfish}]
    A policy profile $\bfs \in \calS_{\policy}^N$ is $(\beta, \epsilon)$-stable if for any $s' \in \calS_{\policy}$, $u \in \bbR^+$, and $i, j \in [N]$,
    \begin{equation}
        \small
        \begin{aligned}
            & \, \bbE[\reward_i(T; s', \bfs_{-j})] \le \bbE[\reward_i(T; \bfs)] - u \\
            \Longrightarrow & \, \bbE[\reward_j(T; s', \bfs_{-j})] \le \bbE[\reward_j(T; \bfs)] + \epsilon - \beta u.
        \end{aligned}
    \end{equation}
\end{definition}
Intuitively, if player $j$ wants to incur a considerable loss $u$ to player $i$, then he will also suffer from a comparable loss of at least $\beta u - \epsilon$.


%% file: paragraphs/method.tex
\section{Proposed Methods} \label{sect:proposed-method}
We first analyze the Nash equilibrium when the arms' expected rewards are known in \sectionref{sect:PNE}. Afterward, based on the equilibrium, we provide the novel Selfish MPMAB with Averaging Allocation (SMAA) approach for players in \sectionref{sect:algorithm}. We theoretically analyze the property of the algorithm in \sectionref{sect:theory}. In these subsections, we assume that each player knows the number of players and is endowed with a different rank (\textit{i.e.}, a unique numbering from $1$ to $N$ for each agent). Finally, in \sectionref{sect:relax}, we relax the assumption and demonstrate that combining the classic Musical Chairs approach \citep{rosenski2016multi} with SMAA could achieve similar theoretical results.

\paragraph{Basic assumption}
Previous works on the zero collision reward setting usually assume that the expected rewards of arms are different \citep{boursier2020selfish,wang2020optimal}. Extending to the shareable arm case, we slightly strengthen the assumption as shown in \assumptionref{assum:unique}.
\begin{assumption} \label{assum:unique}
    For any $k, k' \in [K]$ and $n, n' \in [N]$, we have $\mu_k / n \ne \mu_{k'} / n'$.
\end{assumption}

\begin{remark}
    We further demonstrate the rationality of the assumption by proving that it holds with probability 1 when the expected rewards of different arms $\mu_k$ are sampled randomly  from an absolutely continuous probability distribution with bounded probability densities (such as uniform and beta distributions). Details can be found in \sectionref{sect:justification-assumption}.
\end{remark}

\subsection{Nash Equilibrium When Rewards Are Known} \label{sect:PNE}
We first analyze the Nash equilibrium at each round when players know the expected reward of each arm. Since players are selfish and aim to target their rewards, their strategies at each round will eventually converge to the Nash equilibrium.

At each round, the strategy set for each player is the set of all arms, \textit{i.e.}, $\calS_{\single} = [K]$. For a strategy profile $\bfs \in \calS_{\single}^N$, we use $\bfm(\bfs) = \{m_k(\bfs)\}_{k=1}^K$ to denote the number of players that choose arm $k$. Then the utility of player $j$ is the expected reward at this round, which is given by $\calU_j^{\single}(\bfs) = \mu_{s_j} / m_{s_j}(\bfs)$ according to the averaging allocation mechanism.

Our problem setting at each round is a specific instance of the singleton congestion game (SCG) \citep{ieong2005fast,basat2017game}. Indeed, \citet{ieong2005fast} has developed polynomial-time algorithms to discover the Nash equilibrium that corresponds to the maximal social welfare in SCGs. However, the algorithm is designed for general SCGs, failing to characterizing the detailed properties of our problem. As result, it pose significant challenges to adopt them in online settings where each arms' rewards remain unknown. To address this, we develop the following property for the Nash equilibria in our problem setting, which can be seamlessly integrated within online settings.

\begin{theorem} \label{thrm:PNE}
    Under \assumptionref{assum:unique}, the Nash equilibrium for the game specified by $(\calS_{\single}, \{\calU_j^{\single}\}_{j=1}^N)$ exists. Moreover, for any strategy profile $\bfs$, it is a Nash equilibrium if and only if $\bfm(\bfs) = \bfm^*$ where $\bfm^*$ is given by
    \begin{align}
        \small
        \bfm^* & = \left\{m^*_k =  \floor{\frac{\mu_k}{z^*}} \right\}_{k=1}^K, \label{eq:m-star} \\
        \text{where} \quad z^* & = \sup \left\{z > 0 : h(z) \triangleq \sum_{k=1}^K \floor{\frac{\mu_k}{z}} \ge N\right\}. \label{eq:z-star}
    \end{align}
\end{theorem}
We further use $\calM^* = \{k \in [K]: m_k^* > 0\}$ to represent the arms chosen in the equilibrium.

\begin{remark}
    As shown in \sectionref{sect:justification-strong-pne}, the equilibria described in \theoremref{thrm:PNE} are actually strong Nash equilibria, hence robust to multiple players' deviations.
\end{remark}

\begin{remark}
    Although we only consider pure strategies in \theoremref{thrm:PNE} (\textit{i.e.}, pure Nash equilibrium (PNE)), our evaluation of the randomized strategies in \sectionref{sect:justification-mne} reveals that the total welfare (measured as the total rewards of all players) from any symmetric Mixed Nash equilibrium (MNE) (\textit{i.e.}, all players follow the same randomized strategy) does not exceed that from equilibria in \theoremref{thrm:PNE}. We analyze symmetric MNE due to the computational difficulties inherent in general mixed Nash equilibria \citep{nisan2007algorithmic}, aligning with prior research in algorithm-curated platforms \citep{jagadeesan2022supply}.
\end{remark}

We present an example to better explain the intuition behind \theoremref{thrm:PNE}.

\begin{example} \label{example:PNE}
    Suppose there are $K = 3$ arms with expected rewards $\{\mu_1, \mu_2, \mu_3\} = \{1, 0.4, 0.2\}$ and $N = 3$ players. Then in a Nash equilibrium, two players pull arm $1$ with reward $1$ and one player pulls arm $2$ with reward $0.4$.
\end{example}

\theoremref{thrm:PNE} and \exampleref{example:PNE} demonstrate that when players reach an equilibrium, the number of players that choose each arm is approximately proportional to the corresponding arms' expected rewards. However, this may harm the total welfare (measured as the total rewards of all players). In \exampleref{example:PNE}, the total welfare of all players in the equilibrium is $1.4$. By contrast, if a player deviates from arm $1$ to arm $3$, the total welfare will increase to $1.6$. This phenomenon is well known as the existence of the price of anarchy (PoA) \citep{roughgarden2005selfish,nisan2007algorithmic} in game theory, which indicates the mismatch between agents' utilities and the total welfare.

\paragraph{Analysis of the PoA} We further analyze the bound on the value of PoA. Formally, PoA is measured as the ratio between the maximal total welfare achieved by any strategy profile and the total welfare in the equilibrium.

\begin{proposition} \label{prop:poa}
    PoA is bounded between $1$ and $(N + \min\{K,N\} - 1) / N$. Furthermore, PoA can take value at $1$ and any close to $(N + \min\{K,N\} - 1) / N$.
\end{proposition}


\begin{remark}
    \citet{basat2017game} analyze the PoA of the same setting when $N = 2$ and demonstrate that PoA is at most $1.5$, which is consistent with the proposition when $K \ge 2$. Furthermore, \propositionref{prop:poa} extends their results to more general scenarios where $N$ may exceed $2$. Furthermore, we would like to highlight that \theoremref{thrm:PNE} proves that the Nash equilibrium at each round in our proposed setting is unique with respect to the number of agents on each arm. As a result, all of the Nash equilibria in \theoremref{thrm:PNE} have the same PoA and total social welfare in our setting. Details and further analysis can be found in \sectionref{sect:justification-poa}.
\end{remark}



\subsection{Selfish MPMAB with Averaging Allocation (SMAA)} \label{sect:algorithm}
In this subsection, we present our novel SMAA approach based on the property of the Nash equilibrium demonstrated in \theoremref{thrm:PNE}. Our SMAA approach is inspired by the alternate exploration method \citep{wang2020optimal} while we need to estimate the Nash equilibrium at each round, which becomes more complicated.

\subsubsection{Empirical Statistics}
Consider a player with rank $j$. Let $\tau_{j,k}(t) = \sum_{s=1}^t \bbI[\pi_j(t) = k]$ be the number of rounds that player $j$ chooses arm $k$. The estimation $\hat{\mu}_{j,k}(t)$ of player $j$ on the reward of arm $k$ at round $t$ is given by
\begin{equation}
    \small
    \hat{\mu}_{j,k}(t) = \frac{1}{\tau_{j,k}(t)}\sum_{s=1}^t\bbI[\pi_j(t) = k]X_k(t).
\end{equation}
Exploration is based on the KL-UCB indices \citep{garivier2011kl}. The index of arm $k$ for player $j$ at round $t$ is given by
\begin{equation} \label{eq:kl-ucb}
    \small
    \hat{b}_{j,k}(t) = \sup\left\{q \ge \hat{\mu}_{j,k}(t): \tau_{j,k}(t)\kl(\hat{\mu}_{j,k}(t), q) \le f(t)\right\},
\end{equation}
where $f(t) = \log t + 4 \log (\log t)$.

At round $t$, the estimated equilibrium $\hat{\bfm}_j(t) = \{\hat{m}_{j,k}(t)\}_{k=1}^K$ is calculated by \equationsref{eq:m-star} and \eqref{eq:z-star}
according to the estimated rewards $\{\hat{\mu}_{j,k}(t)\}_{k=1}^K$. In addition, let $\hat{\calM}_j(t) = \{k \in [K]: \hat{m}_{j,k}(t) > 0\}$ be the arms in the estimated equilibrium $\hat{\bfm}_j(t)$. For any arm $k \in \hat{\calM}_j(t)$, let $\hat{r}_{j,k}(t) = \hat{\mu}_{j,k}(t) / \hat{m}_{j,k}(t)$ denote the expected average reward to choose arm $k$ in the estimated equilibrium $\hat{\bfm}_j(t)$ with estimated rewards $\hat{\mu}_{j,k}(t)$.

\subsubsection{Algorithm Details}

\def\NoNumber#1{{\def\alglinenumber##1{}\State #1}\addtocounter{ALG@line}{-1}}
\begin{algorithm}[tb]
    \caption{Selfish MPMAB with Averaging Allocation (SMAA)}
    \label{alg:SMAA}
    \begin{algorithmic}[1]
        \STATE \textbf{Input:} Player rank $j$
        \STATE Let $K' \leftarrow N\cdot\ceil{K/N}$
        \STATE \underline{Initialization phase $1 \le t \le K'$:}
        \STATE Pull each arm at least one time.
        \STATE \underline{Exploration-exploitation phase $t > K'$:}
        \FOR{$t \leftarrow K' + 1$ to $T$}
            \STATE Calculate $\tilde{\tau}_{j,k}(t)$, $\tilde{\mu}_{j,k}(t)$, $\tilde{b}_{j,k}(t)$, $\tilde{\calM}_j(t)$, $\tilde{\bfm}_j(t)$, $\tilde{r}_{j,k}(t)$ according to \equationref{eq:block-estimation}
            \STATE Calculate the arms chosen in this block $\tilde{l}_j(t)$ according to \equationref{eq:sequence-in-block}
            \STATE Calculate the index $i \leftarrow (t + j) \mo N + 1$
            \STATE Let $k$ be the $i$-th element in $\tilde{l}_j(t)$
            \IF{$i = N$}
                \STATE Calculate $\calH_j$ according to \equationref{eq:calH}
                \IF{$\calH_j(t) = \emptyset$}
                    \STATE $\pi_j(t) \leftarrow k$
                \ELSE
                    \STATE With probability $\frac{1}{2}$, $\pi_j(t) \leftarrow k$
                    \STATE With probability $\frac{1}{2}$, $\pi_j(t) \leftarrow k'$ chosen uniformly at random in $\calH_j(t)$
                \ENDIF
            \ELSE
                \STATE $\pi_j(t) \leftarrow k$
            \ENDIF
            \STATE Pull arm $\pi_j(t)$
        \ENDFOR
    \end{algorithmic}
\end{algorithm}

The pseudo-code is shown in \algorithmref{alg:SMAA} and the algorithm consists of two phases.

\paragraph{Initialization phase} Each player first conducts an initialization phase with $K' = N \ceil{K / N}$ rounds. Since $K' \ge K$ by construction, each player can pull each arm at least one time. We set the rounds for initialization as $K'$ instead of $K$ because $K' \equiv 0 (\mo N)$, which matches the second exploration-exploitation phase that is conducted in blocks with $N$ rounds each.

\paragraph{Exploration-exploitation phase} Afterward, each player enters the exploration-exploitation phase. The whole procedure is conducted in blocks with $N$ rounds each. Generally speaking, in each block, each player sequentially pulls the arms in the estimated Nash equilibrium based on his historical observations unless the player pulls the arm with the smallest average reward for the last time. In this case, he will explore other arms with probability $1/2$.

Formally, the behaviors of player $j$ in a block are based on the estimations up to the end of the last block. As a result, the policy at round $t$ is determined by the following estimations (with the tilde notation).
\begin{equation} \label{eq:block-estimation}
    \begin{small}
    \begin{aligned}
        & \tilde{\tau}_{j,k}(t) = \tau_{j,k}\left(\floor{\frac{t-1}{N}}N\right), \tilde{\mu}_{j,k}(t) = \hat{\mu}_{j,k}\left(\floor{\frac{t-1}{N}}N\right), \\
        & \tilde{b}_{j,k}(t) = \hat{b}_{j,k}\left(\floor{\frac{t-1}{N}}N\right), \tilde{\calM}_j(t) = \hat{\calM}_j \left(\floor{\frac{t-1}{N}}N\right), \\
        & \tilde{\bfm}_j(t) = \hat{\bfm}_j \left(\floor{\frac{t-1}{N}}N\right), \tilde{r}_{j,k}(t) = \tilde{r}_{j,k} \left(\floor{\frac{t-1}{N}}N\right).
    \end{aligned}
    \end{small}
\end{equation}
Here the value of $\floor{(t - 1) / N}$ is the last round in the previous block.

For each round $t$, player $j$ first calculates the list of arms to choose in the corresponding block as shown in Line 7-8 in \algorithmref{alg:SMAA}. In particular, player $j$ sorts the arms in the estimated equilibrium $\tilde{\calM}_j(t)$ according to the average reward $\tilde{r}_{j,k}(t)$ by descending order and get the indices of the arms $k_1, k_2, \dots, k_{|\tilde{\calM}_j(t)|}$. Then he sequentially aligns these arms and each arm $k_i$ is further repeated by $\tilde{m}_{j,k_i}$ times. Formally, the arms chosen in the corresponding block are given by
\begin{equation} \label{eq:sequence-in-block}
    \small
    \tilde{l}_j(t) = \left(\underbrace{k_1, \dots, k_1}_{\tilde{m}_{j, k_1}(t) \text{ times}}, \dots, \underbrace{k_{a}, \cdots, k_{a}}_{\tilde{m}_{j, k_{a}}(t) \text{ times}}\right),
\end{equation}
where $a = |\tilde{\calM}_j(t)|$.
We demonstrate how $\tilde{l}_j(t)$ is calculated through the numerical showcase provided in \exampleref{example:PNE}.
\begin{example}
    Suppose the estimated expected reward of player $j$ at a round $t$ is $\{\tilde{\mu}_{j,k}(t)\}_{k=1}^K=\{1, 0.4, 0.2\}$. Then the estimated number of players on each arm in the equilibrium $\{\tilde{m}_{j,k}(t)\}_{k=1}^K=\{2, 1, 0\}$, and we have $\tilde{\calM}_j(t) = \{1, 2\}$. As a result, the average reward $\tilde{r}_{j,k}(t)$ for arms $k$ in $\tilde{\calM}_j(t)$ is $0.5$ for arm $1$ and $0.4$ for arm $2$. After sorting the arms in $\tilde{\calM}_j(t)$ according to $\tilde{r}_{j,k}(t)$ by descending order and repeating each arm with $\tilde{m}_{j,k}(t)$ times, we have $\tilde{l}_j(t) = \{1, 1, 2\}$.
\end{example}

Afterward, as shown in Line 9-10, player $j$ chooses the element $k$ in $\tilde{l}_j(t)$ with index $i = (t + j) \mo N + 1$ as a candidate to pull at this round. Moreover, when the player chooses the last element in $\tilde{l}_j(t)$ (\textit{i.e.}, $i = N$), as depicted in Line 13-18, he will explore the arms in $\calH_j(t)$ uniformly with probability $1/2$ and the set $\calH_j(t)$ is defined as
\begin{equation} \label{eq:calH}
    \small
    \calH_j(t) = \left\{k' \not\in \tilde{\calM}_j(t): \tilde{b}_{j,k'}(t) \ge \tilde{r}_{j,k_a}\right\}.
\end{equation}
Otherwise, the player will pull arm $k$ as shown in Line 20.

\subsection{Theoretical Analysis of SMAA} \label{sect:theory}
\subsubsection{Performances} \label{sect:theory-ours}
Let
\begin{equation} \label{eq:delta-0}
    \small
    \delta_0 = \min_{x, y \in \Delta, x \ne y} |x - y|,
\end{equation}
where $\Delta = \{\mu_k / n: k \in [K], n \in [N]\} \cup \{0\}$ is the set of all possible average rewards. Under \assumptionref{assum:unique}, we have $\delta_0 > 0$. Now we present the theoretical results about \algorithmref{alg:SMAA} on the metrics introduced in \sectionref{sect:metrics}.

\begin{theorem} \label{thrm:main-regret}
    Under \assumptionref{assum:unique}, suppose $0 < \delta < \delta_0 / 2$. Then when all players follow \algorithmref{alg:SMAA}, the expected regret for each player $j$ is upper-bounded by
    \begin{equation}
        \small
        \begin{aligned}
            \bbE\left[\regret_j(T)\right] \le & \, \sum_{k \not\in \calM^*} \frac{(z^* - \mu_k)(\log T + 4 \log (\log T))}{\kl(\mu_k + \delta, z^* - \delta)} \\
            & \, + 10N^3K(13K + \delta^{-2}).
        \end{aligned}
    \end{equation}
\end{theorem}
\begin{remark}
    The proof is provided in \appendixref{sect:proof-main-regret}. We highlight that $\delta_0$ and $\delta$ is unknown to the algorithm though the bounds depend on $\delta$. This theorem demonstrates that each player can effectively maximize their own reward since the regret is sublinear. In addition, by letting $T$ tend to $\infty$ and then $\delta$ tend to $0$, we can get that
    \begin{equation}
        \small
        \limsup_{T \rightarrow \infty} \frac{\bbE\left[\regret_j(T)\right]}{\log T} \le \sum_{k \not\in \calM^*} \frac{z^* - \mu_k}{\kl(\mu_k, z^*)},
    \end{equation}
    which is asymptotically optimal and matches the lower bound in \theoremref{thrm:lower-bound} in \sectionref{sect:lower-bound} when reward distributions are Bernoulli.
\end{remark}

\begin{remark}
    It is also reasonable to define the regret as the gap between each agent's total reward and the average reward in the Nash equilibrium as shown in \theoremref{thrm:PNE}. We demonstrate in \sectionref{sect:justification-another-regret} that our algorithm still achieves a $O(\log T)$ regret under this formulation.
\end{remark}

\begin{theorem} \label{thrm:main-noneq}
    Under \assumptionref{assum:unique}, suppose $0 < \delta < \delta_0 / 2$. Then when all players follow \algorithmref{alg:SMAA}, the expected number of non-equilibrium rounds demonstrated in \theoremref{thrm:PNE} is upper bounded by
    \begin{equation}
        \small
        \begin{aligned}
            \bbE[\noneq(T)] \le & \, N\sum_{k \not\in \calM^*} \frac{\log T + 4 \log (\log T)}{\kl(\mu_k + \delta, z^* - \delta)} \\
            & \, + 10N^3K(13K + \delta^{-2}).
        \end{aligned}
    \end{equation}
\end{theorem}
\begin{remark}
    The proof is provided in \appendixref{sect:proof-main-noneq}. This theorem demonstrates that when all players follow \algorithmref{alg:SMAA}, they will eventually converge to the Nash equilibrium since the expected number of non-equilibrium rounds is sublinear. Although it is well known that no internal regret strategies converge in average to the set of correlated equilibria \citep{nisan2007algorithmic}, in this paper we prove that our algorithm can converge to the pure Nash equilibrium at each round. As a result, standard no internal regret strategies cannot be adopted directly in our problem setting.
\end{remark}

\begin{theorem} \label{thrm:selfish}
    Under \assumptionref{assum:unique}, suppose $0 < \delta < \delta_0 / 2$. Then the policy profile where all players follow \algorithmref{alg:SMAA} is an $\epsilon$-Nash equilibrium \textit{w.r.t.} the game specified by $(\calS_{\policy}, \{\calU_j^{\policy}\}_{j=1}^N)$ and is $(\beta, \epsilon + \beta\gamma)$-stable with
    \begin{equation}
        \small
        \begin{aligned}
            \beta = & \, \frac{\delta_0}{z^*}, \\
            \epsilon = & \, \sum_{k \not\in \calM^*} \frac{(z^* - \mu_k)(\log T + 4 \log (\log T))}{\kl(\mu_k + \delta, z^* - \delta)} \\
            & \, + 10N^3K(13K + \delta^{-2}), \\
            \gamma = & \, \sum_{k \not\in \calM^*}\frac{\log T + 4\log(\log T)}{\kl(\mu_k + \delta, z^* - \delta)} + 10N^3K(13K + \delta^{-2}).
        \end{aligned}
    \end{equation}
    Here $z^*$ is provided in \equationref{eq:z-star}.
\end{theorem}

\begin{remark}
    The proof is provided in \appendixref{sect:proof-selfish}. This theorem demonstrates that any player can not significantly improve his reward by deviation since the policy profile is $\epsilon$-Nash equilibrium with $\epsilon = O(\log T)$. In addition, if any selfish player wants to harm another player's reward significantly (such as $O(T)$), then he will also suffer from a big loss (also $O(T)$). We also note that the bound can be tighter ($\epsilon, \gamma$ can be smaller) in reality since we prove the theorem by assuming the strategic player knows all the parameters beforehand. As a result, SMAA is more stable to the strategic deviation of a single player in reality since the actual $\epsilon$ and $\gamma$ are smaller than those demonstrated in the theorem.
\end{remark}

\subsubsection{Regret Lower Bound} \label{sect:lower-bound}
We now present the regret lower bound for each player. Similar to the previous works on KL-UCB \citep{garivier2011kl,combes2015combinatorial,besson2018multi}, we focus on the Bernoulli reward cases. Formally, we use $\calI_{N,K}^{\ber} \subseteq \Xi^K$ to denote the set of all problem instances $(\xi_1, \xi_2, \dots, \xi_K)$ that satisfies \assumptionref{assum:unique} and each $\xi_k$ is a Bernoulli distribution. 


\begin{definition} \label{defn:consistent}
    We say a policy profile $\bfA = (A_1, A_2, \dots, A_N) \in \calS_{\policy}^N$ is \textit{consistent} if for any problem instance $(\xi_1, \xi_2, \dots, \xi_K) \in \calI_{N, K}^{\ber}$ with expectations $\mu_1, \mu_2, \dots, \mu_K$ and any $k \in \calM^*, j \in [N]$,
    \begin{equation}
        \small
        \forall \alpha \in (0, 1], \quad \frac{m_k^*}{N}T - \bbE[\tau_{j,k}(T)] \le o(T^\alpha).
    \end{equation}
    Here $\tau_{j,k}(T)$ is the number of times that arm $k$ is chosen. $m_k^*$ is the number of times arm $k$ is chosen in the equilibrium calculated according to \theoremref{thrm:PNE}. $\calM^*$ is the set of arms that in the equilibrium.
\end{definition}

We note that \algorithmref{alg:SMAA} satisfies the consistent condition since players sequentially choose the arms in the Nash equilibrium and sub-optimal arms are chosen at most $O(\log T)$ times (demonstrated by \lemmasref{lemma:subseteq}, \ref{lemma:subset-expectation}, and \ref{lemma:bad-arms} in \appendixref{sect:proof-theory}).
Any policy profile $\bfA$ that is not consistent cannot achieve a desirable regret guarantee, when $\bfA$ satisfies a fairness condition that all players are expected to pull each arm for similar times. The formal formulation and the proof of this claim can be found in \sectionref{sect:justification-consistent}. Note that this fairness condition is also mentioned by \citep{besson2018multi} and is satisfied by all policy profiles that assigning the same algorithm to all players. Both \algorithmref{alg:SMAA} and \algorithmref{alg:SMAA-relaxed} satisify the fairness condition.

With this definition, we show a problem-dependent asymptotic lower bound for consistent algorithms that only use the arms’ reward information.

\begin{theorem} \label{thrm:lower-bound}
    Suppose a policy profile $\bfA \in \calS_{\policy}^N$ is consistent as defined in \definitionref{defn:consistent}. Assume $\bfA$ only uses arms' reward information. Formally, for each agent $j$, $\pi_j(t)$ is $\calF_{j}(t)$-measurable and 
    \begin{equation}
        \begin{aligned}
            \calF_{j}(t) = \sigma( & X_{\pi_j(1)}(1), X_{\pi_j(2)}(2), \dots, X_{\pi_j(t - 1)}(t - 1),  \\
            & U_0, U_1, \dots, U_{t}).
        \end{aligned}
    \end{equation}
    Here $\sigma(\cdot)$ denote the sigma algebra and $U_0$, $U_1$, \dots, $U_{t}$ denote the external sources of randomness to determine $\pi_j(t)$ (\textit{e.g.}, the randomness in Lines 16-17 in \algorithmref{alg:SMAA}).
    Then for any problem instance $\bfxi = (\xi_1, \xi_2, \dots, \xi_K) \in \calI_{N, K}^{\ber}$ with expectations $\mu_1, \mu_2, \dots, \mu_K$, we have
    \begin{equation} \label{eq:tau-lower-bound}
        \small
        \forall k \not\in \calM^*, \quad \liminf_{T \rightarrow \infty} \frac{\bbE[\tau_{j,k}(T)]}{\log T} \ge \frac{1}{\kl(\mu_k, z^*)}.
    \end{equation}
    Furthermore, we have
    \begin{equation}
        \small
        \forall j \in [N], \quad \liminf_{T \rightarrow \infty} \frac{\bbE[\regret_j(T)]}{\log T} \ge \sum_{k \not\in \calM^*}\frac{z^* - \mu_k}{\kl(\mu_k, z^*)}.
    \end{equation}
\end{theorem}

\begin{remark}
    The proof is provided in \appendixref{sect:proof-lower-bound}. Here we additionally assume that algorithms only use the arms' reward information to avoid the effect of a strategic player that deliberately collides with other players. \algorithmref{alg:SMAA} satisfies the condition. Therefore, under the conditions, our algorithm's regret matches the lower bound asymptotically.
\end{remark}

\subsection{SMAA without Knowledge of $N$ and Rank} \label{sect:relax}
\begin{figure*}
    \centering
    \includegraphics[width=0.95\linewidth]{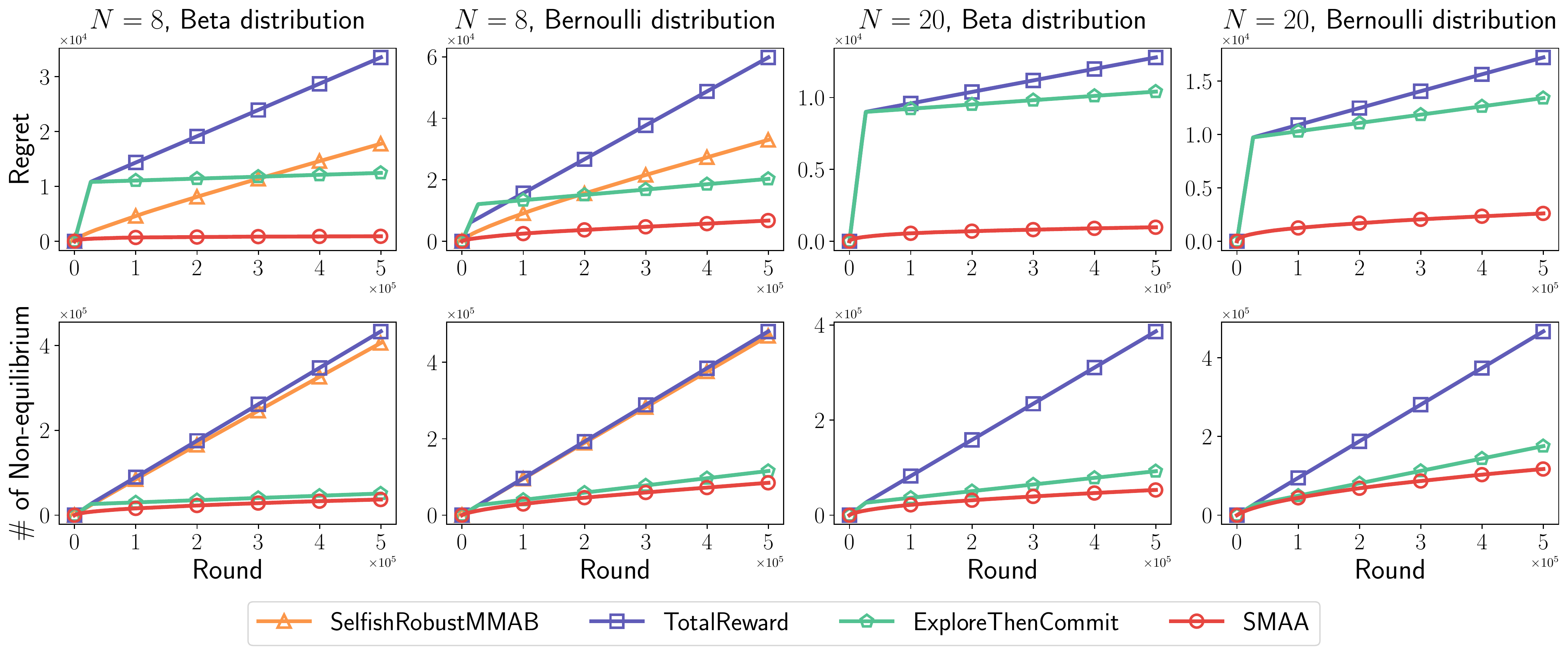}
    \vspace{-15px}
    \caption{Regret and number of non-equilibrium rounds curve when $N$ and rank is known. $K$ is set to $10$. We note that SelfishRobustMMAB \citep{boursier2020selfish} considers the non-shareable arm setting and can only be applied to the settings when $N \le K$.}
    \vspace{-15px}
    \label{fig:all}
\end{figure*}

In previous subsections, we assume that players have the knowledge of the total number of players $N$ and each player has a different rank. However, this information may not be available in real-world scenarios. In this case, we further assume that $N \le K$ and there is an indicator to inform each player whether he meets a collision, which is common in previous works \citep{rosenski2016multi,boursier2019sic,bande2019multi}.

With these assumptions, we could adopt the Musical Chairs approach \citep{rosenski2016multi} to estimate the number of players and assign each player a different rank. The details of the algorithm is provided in \algorithmref{alg:SMAA-relaxed} in  \appendixref{sect:relaxed-setting}.
We demonstrate that the Musical Chairs approach is robust to a single player's deviation and  prove similar results for this algorithm with the conclusions in \sectionref{sect:theory-ours}.

\begin{corollary} [Informal version of \theoremref{thrm:relaxed-setting}] \label{coro:main-relaxed}
    Suppose \assumptionref{assum:unique} hold. Then \algorithmref{alg:SMAA-relaxed} satisfies that $\forall j \in [N], \bbE[\regret_j(T)] \le O(\log T)$. $\bbE[\noneq(T)] \le O(\log T)$. In addition, the policy profile where all players follow \algorithmref{alg:SMAA-relaxed} is an $\epsilon$-Nash equilibrium and is $(\beta, \epsilon)$-stable with $\epsilon = O(\log T)$ and $\beta = \delta_0 / z^*$.
\end{corollary}

%% file: paragraphs/experiments.tex
\section{Synthetic Experiments} \label{sect:experiments}

\begin{figure}[t]
    \centering
    \includegraphics[width=\linewidth]{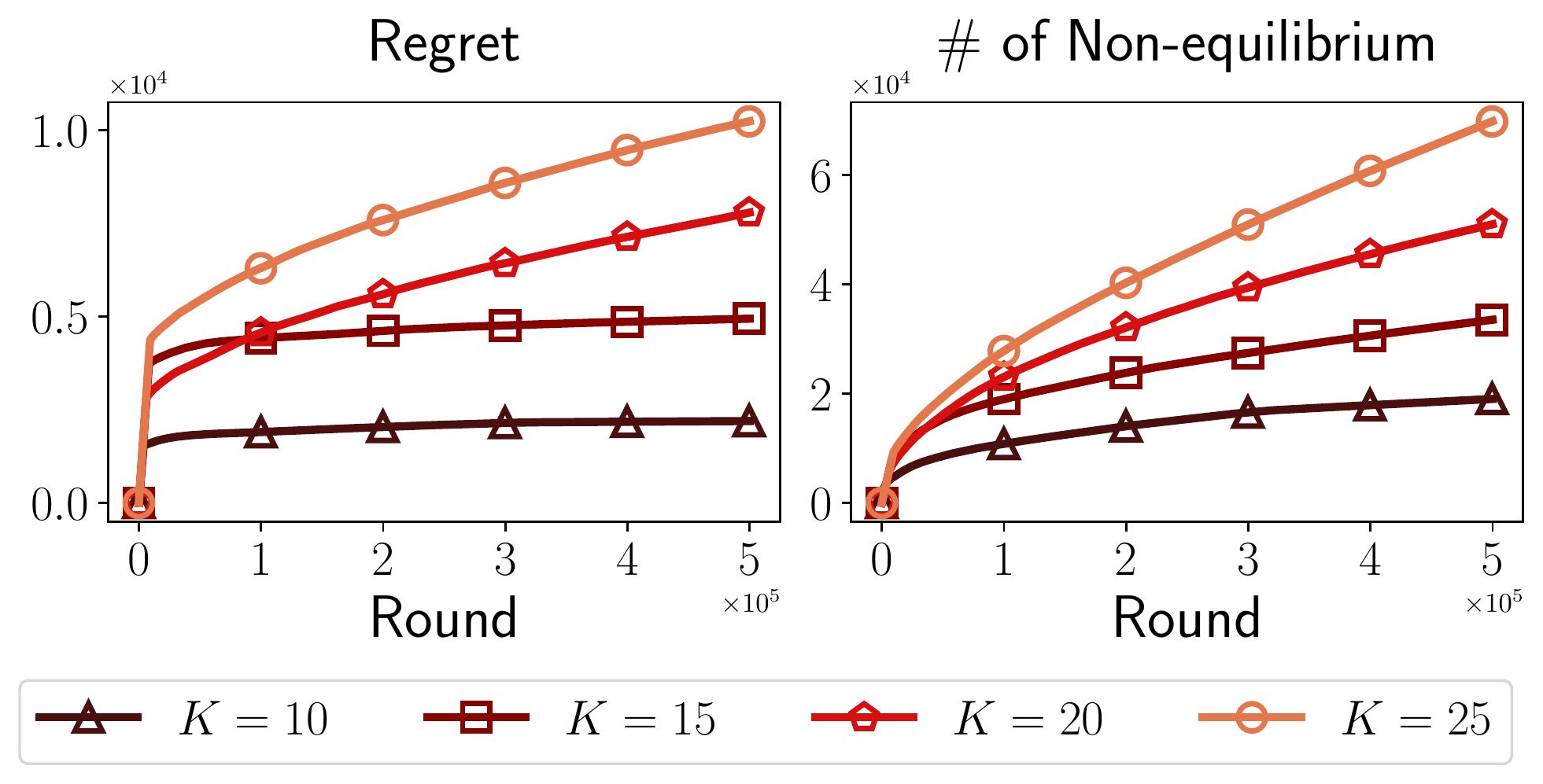}
    \vspace{-25px}
    \caption{Regret and the number of non-equilibrium rounds curve when $N$ and rank are unknown. Here $N = 8$, $T = 500,000$, and the reward of each arm follows the beta distribution.}
    \label{fig:relaxed}
    \vspace{-10px}
\end{figure}

We validate the effectiveness of the proposed SMAA method through synthetic experiments. The source code is available at \url{https://github.com/windxrz/SMAA}.

\subsection{Experiments When $N$ and Rank Are Known}
We first conduct experiments in a setting where players know $N$ and are endowed with different ranks.

\paragraph{Data-generating process}
In this setting, we set the number of arms as $K=10$ and the number of players $N \in \{8, 20\}$ to test the models' performance in both $K < N$ and $K \ge N$ scenarios. The number of total rounds $T$ is set to $500,000$.
In addition, for each arm's reward distribution,  we consider two common distributions supported on $[0, 1]$, including the beta distribution and Bernoulli distribution. For the beta distribution $\text{Beta}(\alpha, \beta)$, we randomly sample the two shape parameters $\alpha$ and $\beta$ uniformly in $[0,5]$ for each arm. For the Bernoulli distribution $\text{Ber}(p)$, we randomly sample the probability parameter $p$ uniformly in $[0,1]$.

\paragraph{Baselines}
We implement the following baselines. \textbf{TotalReward} \citep{bande2019multi} considers the shareable reward case while its target is to maximize the total reward of all players. \textbf{SelfishRobustMMAB} \citep{boursier2020selfish} assumes players can be selfish while the arms are not shareable and players get no reward upon collision. \textbf{ExploreThenCommit} first randomly explores the rewards of arms and then commits to the Nash equilibrium based on the estimation in the exploration phase. Details of these baselines are provided in \appendixref{sect:app-exp}.

\paragraph{Analysis}
We implement the experiments for $100$ different simulations by resampling the arms' reward distributions. We then evaluate our method (SMAA) and baselines by calculating the average regret among all players and the number of non-equilibrium rounds according to \equationsref{eq:regret} and~\eqref{eq:non-equ}.  The results are shown in \figureref{fig:all} and our method outperforms all baselines in all settings with various data-generating processes.

On the one hand, the TotalReward and SelfishRobustMMAB baselines consider the different MPMAB scenarios and they can not directly be applied here for the selfish player setting with the averaging collision model. On the other hand, compared with the explore-then-exploit-based baseline ExploreThenCommit, our method shows a better exploration-exploitation trade-off, leading to smaller regret and non-equilibrium rounds.

\subsection{Experiments When $N$ and Rank Are Unknown}
We then test under the setting when players do not know $N$ and ranks.

\paragraph{Data-generating process}
In this setting, we fix the number of players as $N = 8$, and the total rounds as $T = 500,000$. In addition, we simulate each arm's reward using the beta distribution. We set $K \in \{10, 15, 20, 25\}$ to test the convergence and performance of our approach.

\paragraph{Analysis}
Analogous to the experiments when $N$ and rank are known, we implement the experiments for $100$ different simulations by resampling the arms' reward distributions. We then evaluate SMAA by calculating the average regret among all players and the number of non-equilibrium rounds. As shown in~\figureref{fig:relaxed}, we observe that our method could effectively converge to Nash equilibrium and achieve small regret since the slopes of these curves become smoother as the number of rounds increases. In addition, we observe from \figureref{fig:relaxed} that when the number of arms $K$ becomes larger, the problem becomes more challenging, and both the regrets and the number of non-equilibrium rounds increase. However, our method could still effectively converge since the proportion of non-equilibrium rounds is about $14\%$ ($69,959 / 500,000$) when $K=25$ at round $500,000$, and the proportion is still decreasing due to the declining slope of the curve.

%% file: paragraphs/related_works.tex
\section{Related Works} \label{sect:related-works}
\paragraph{Multi-player multi-armed bandit}
Multi-armed bandit (MAB) problems \citep{lai1985asymptotically,bubeck2012regret,slivkins2019introduction,lattimore2020bandit} have been extensively studied in different online settings \citep{kveton2015cascading,lattimore2016causal,agrawal2016linear,xu2022product,ferreira2022learning}. Extending MAB, multi-player multi-armed bandit (MPMAB) \citep{liu2008restless,jouini2009multi,jouini2010upper,branzei2021multiplayer} considers the problem where multiple players act on a single multi-armed bandits problem instance \citep{boursier2022survey}. Most papers in MPMAB assume that players cannot get any reward when a collision occurs \citep{rosenski2016multi,besson2018multi,boursier2019sic,boursier2020selfish,shi2020decentralized,shi2021heterogeneous,huang2022towards,lugosi2022multiplayer}. A thorough survey can be found in \citep{boursier2022survey}.

\paragraph{Different collision models in MPMAB}
Several papers assume different reward allocation mechanisms when a collision occurs.
\citet{liu2020competing,liu2021bandit,jagadeesan2021learning,basu2021beyond} assume that the player with the maximal expected reward would get the reward.
Some works assume a threshold \citep{bande2019multi,youssef2021resource,bande2021dynamic,magesh2021decentralized} or capacity \citep{wang2022multi,wang2022multiple} for each arm.
\citet{shi2021multi,pacchiano2021instance} assume that each player gets a reduced reward when a collision occurs.
\citet{boyarski2021distributed} focus on a heterogeneous reward setting. Our work differs from these works mainly in that we assume an averaging allocation mechanism with homogeneous arms' rewards. In addition, most of these works do not consider selfish player behaviors and the target is to maximize the total welfare.

\paragraph{Strategic players in MPMAB}
There are two kinds of strategic behaviors in MPMAB, including jammers and selfish players. Jammers \citep{attar2012survey,wang2015jamming,sawant2018distributed,sawant2019learning} aim to perturb the cooperative players as much as possible even at the cost of their rewards. Several works also consider the selfish behaviors in MPMAB while they assume that at most one player will get the reward. Specifically, \citet{boursier2020selfish} considers the standard zero-reward collision model. \citet{liu2020competing,liu2021bandit,jagadeesan2021learning} consider the stable matching setting.
Compared to these works, we assume that arms are shareable.

\paragraph{Game analysis in applications with shareable resources}
Ranging from algorithm-curated platforms \citep{hron2022modeling,jagadeesan2022supply,ben2019recommendation,yao2022learning,yao2022learningb,yao2023bad} to cognitive radio communications \citep{riahi2019game,riahi2016optimization}, game theory plays an important role in allocating limited resources among multiple strategic players. For algorithm-curated platforms, the equilibrium of content providers is analyzed \citep{basat2017game,ben2018game,ben2019recommendation,jagadeesan2022supply,hron2022modeling}. Regarding cognitive radio communications, \textit{i.e.}, another commonly analyzed strategic scenario, \citet{riahi2016optimization,riahi2019game} discuss various types of equilibrium among multiple strategic transmitters. However, most of these works analyze the equilibrium only and do not analyze the dynamics of players. Although this gap has been overcome by several recent works~\citep{riahi2019game,basat2017game,ben2019convergence,ben2020content}, they still assume that players know the reward for each resource.

In addition, contrary to \citet{yao2023bad}'s analysis of the Price of Anarchy (PoA) in coarse correlated equilibria (CCE) for top-$K$ recommendations, our study situates within a MPMAB framework, concentrating on pure Nash equilibrium (PNE). Our aim is to develop algorithms that maximize individual agent revenue and consistently converge to the PNE, diverging from \citet{yao2023bad} focus on CCE's PoA and the assumption of no-regret dynamics.



%% file: paragraphs/conclusion.tex
\section{Conclusion} \label{sect:conclusion}
In this paper, we propose a novel multi-player multi-armed setting with an averaging allocation model to characterize the selfish behaviors of players in applications with limited and shareable resources. To design policies for each player in this setting, we first analyze the Nash equilibrium when players know the expected reward of each arm. Based on the equilibrium, we further propose a novel Selfish MPMAB with Averaging Allocation (SMAA) approach for each player. We theoretically demonstrate that SMAA could achieve a good regret guarantee for each player when all players follow the algorithm and it is robust to a single player's strategic deviation.

\section*{Acknowledgment}
Peng Cui's research was supported in part by National
Natural Science Foundation of China (No. U1936219, 62141607), and National Key R\&D Program
of China (No. 2018AAA0102004). Bo Li’s research was supported by the National Natural Science Foundation of China (No.72171131, 72133002); the Technology and Innovation Major Project of the Ministry of Science and Technology of China under Grants 2020AAA0108400 and 2020AAA0108403. We would like to thank Zi Qian and anonymous reviewers for the helpful feedback.

%% file: paragraphs/appendix-additional.tex
\section{Additional Theoretical Results} \label{sect:additional}
\subsection{Justification of \assumptionref{assum:unique}} \label{sect:justification-assumption}
\begin{proposition} \label{prop:justification-assumption}
    Suppose the expected rewards $\mu_1, \mu_2, \dots, \mu_K \in (0, 1]$ of arms are \textit{i.i.d.} and drawn from a absolutely continuous probability distribution $D$ with bounded probability densities $p(\mu)$ (\textit{i.e.}, for all $\mu \in (0, 1]$, $p(\mu) \le M$ for a constant $M$). Then \assumptionref{assum:unique} holds with probability $1$.
\end{proposition}

The proof is provided in \sectionref{sect:proof-assumption}.

\subsection{Extending the Single-Agent Deviation Setting to Multiple-Agent Deviation Setting} \label{sect:justification-strong-pne}
We prove that every Nash equilibrium in \theoremref{thrm:PNE} is a strong Nash equilibrium.
\begin{proposition} \label{prop:strong-pne}
    For a strategy profile $\bfs$, we say strategies $\bfs_B' \in \prod_{j \in B} \calS_{\single}$ are a beneficial deviation for a subset $B$ of players if
    \begin{equation}
        \forall j \in B, \quad \calU^{\single}_j(\bfs_B', \bfs_{-B}) \ge \calU^{\single}_j(\bfs)
    \end{equation}
    with the inequality holding strictly for at least one player of $B$. Then every Nash equilibrium $\bfs$ in \theoremref{thrm:PNE} has no coalition of players with a beneficial deviation.
\end{proposition}

The proof is provided in \sectionref{sect:proof-strong-pne}.

\subsection{Randomized Strategy at Each Round} \label{sect:justification-mne}
At each round, the strategy set for each player is the set of all arms, \textit{i.e.}, $\calS_{\single} = [K]$. Now suppose player $j$ follows randomized strategy $\bfsigma_j = (\sigma_{j, 1}, \sigma_{j,2}, \dots, \sigma_{j,K})$ on $\calS_{\single}$ where $\sigma_{j,k}$ represents the probability for player $j$ to choose arm $k$. Let $\bfsigma = \bfsigma_1 \times \bfsigma_2 \times \dots \times \bfsigma_N$ be the strategy profile of all players. Then the social welfare under a mixed strategy is given by
\begin{equation}
    W(\bfsigma) = \bbE_{\bfs \sim \bfsigma}\left[\sum_{k=1}^K \mu_k \bbI[m_k(\bfs) > 0]\right] = \sum_{k = 1}^K \mu_k \bbP_{\bfs \sim \bfsigma}[m_k(\bfs)> 0].
\end{equation}
The welfare under the pure Nash equilibrium \theoremref{thrm:PNE} is given by
\begin{equation}
    W^{\text{PNE}} = \sum_{k \in \calM^*} \mu_k.
\end{equation}

\begin{definition} [Mixed Nash equilibrium]
    The randomized strategy profile $\bfsigma = \bfsigma_1 \times \bfsigma_2 \times \dots \times \bfsigma_N$ is a mixed Nash equilibrium if for every player $j \in [N]$ and every unilateral deviation $s_j'$, we have
    \begin{equation}
        \bbE_{\bfs \in \bfsigma}[\calU_j(\bfs)] \ge \bbE_{\bfs \in \bfsigma}[\calU_j(s_j', \bfs_{-j})].
    \end{equation}
\end{definition}

We provide the following property to show that any symmetric mixed Nash equilibrium's total welfare is not larger than the welfare obtained by the pure Nash equilibrium at each round.
\begin{proposition} \label{prop:mne-vs-pne}
    Suppose $\bfsigma = \bfsigma_1 \times \bfsigma_2 \times \dots \times \bfsigma_N$ is a symmetric mixed Nash equilibrium (\textit{i.e.}, $\bfsigma_1 = \bfsigma_2 = \dots = \bfsigma_N$). Then
    \begin{equation}
        W(\bfsigma) \le W^{\text{PNE}}.
    \end{equation}
\end{proposition}

The proof is provided in \sectionref{sect:proof-mne-vs-pne}. We further find that it is actually possible that players pull arms that is not in $\calM^*$ when all players follow a symmetric mixed Nash equilibrium. However, the welfare under the symmetric mixed Nash equilibrium is not larger than the welfare under the pure Nash equilibrium as shown by \propositionref{prop:mne-vs-pne}.
\begin{example}
    Set $N = K = 3$ and $\mu_1, \mu_2, \mu_3 = 1, 0.6, 0.48$. By simulation, we have the profile $\bfsigma$ such that
    \begin{equation}
        \forall j \in [K], \quad \sigma_{j,1}\approx 0.705, \sigma_{j,2}\approx 0.254, \sigma_{j,3} \approx 0.041
    \end{equation}
    formulates a symmetric mixed Nash equilibrium. We can verify that all arms are possible to be chosen in the mixed Nash equilibrium while the social welfare is given by
    \begin{equation}
        W(\bfsigma) = \bbE_{\bfs \in \bfsigma}\left[\sum_{j=1}^N\calU_j^{\single}(\bfs)\right] \approx 1.382.
    \end{equation}
    In addition, we have $\calM^* = \{1, 2\}$ and the welfare for the pure Nash equilibrium is
    \begin{equation}
        W^{\text{PNE}} = 1 + 0.6 = 1.6.
    \end{equation}
\end{example}

\subsection{Formal Analysis of the bound on PoA} \label{sect:justification-poa}
Let $\calC^*$ be the set of the top $\min\{N,K\}$ arms with the highest expected rewards. Then the total welfare under the equilibrium demonstrated in \theoremref{thrm:PNE} is given by
\begin{equation}
    W^{\text{PNE}} = \sum_{k \in \mathcal{M}^*} \mu_k
\end{equation}
while the total welfare for maximal social welfare case is given by
\begin{equation}
    W^{\text{MAX}} = \sum_{k \in \calC^*} \mu_k.
\end{equation}
PoA is measured as follows
\begin{equation}
    \poa = W^{\text{MAX}} / W^{\text{PNE}}.
\end{equation}

\begin{proposition} [Restatement of \propositionref{prop:poa}] \label{prop:poa-restatement}
    PoA is bounded between $1$ and $(N + \min\{K,N\} - 1) / N$. Furthermore, PoA can take value at $1$ and any close to $(N + \min\{K,N\} - 1) / N$.
\end{proposition}

The proof of \propositionref{prop:poa} can be found in \sectionref{sect:proof-poa}. We further analyze the effect of $N$ and $K$ on the upper bound.
\begin{itemize}
    \item The upper bound increases \textit{w.r.t.} $K$ when $K < N$ while remains unchanged when $K \ge N$. This is because only at most $N$ arms with highest expected rewards are in the set $\calC^*$ while remaining arms do not affect the value of PoA.
    \item The upper bound increases \textit{w.r.t.} $N$ when $N < K$ and decreases when $N \ge K$. This is because when $N < K$, the difference between $W^{\text{MAX}}$ and $W^{\text{PNE}}$ can be $(N-1)/N \cdot W^{\text{PNE}}$ and this difference increases \textit{w.r.t.} $N$. In addition, when $N \ge K$, it is more easier to make all arms in the equilibrium set $\mathcal{M}^*$ when $N$ is large. Even when some arms in $\calC^*$ are not in the equilibrium, the number of such arms are bounded by $K$, making them not significantly affect the PoA value.
\end{itemize}

\subsection{Another Regret Formulation and Regret Bound Analysis} \label{sect:justification-another-regret}
Consider an alternative regret that compares each agent's total reward with the average reward in the Nash equilibrium. Formally, it is given by the follows
\begin{equation} \label{eq:regret-new}
    \regret_j'(T) = \sum_{t=1}^T \left(r^* - R_j(t)\right),
\end{equation}
where
\begin{equation}
    r^* = \frac{1}{N}\sum_{k \in \calM^*} \mu_k.
\end{equation}
We can provide the regret bound \textit{w.r.t.} this regret by the following proposition.
\begin{proposition} \label{prop:regret-new}
    Under \assumptionref{assum:unique}, suppose $0 < \delta < \delta_0 / 2$. Then when all players follow \algorithmref{alg:SMAA}, the expected regret for each player $j$ is upper-bounded by
    \begin{equation}
        \bbE\left[\regret_j'(T)\right] \le \sum_{k \not\in \calM^*} \frac{(z^* - \mu_k)(\log T + 4 \log (\log T))}{\kl(\mu_k + \delta, z^* - \delta)} + O(1).
    \end{equation}
\end{proposition}

The proof is provided in \sectionref{sect:proof-regret-new}.

\subsection{Justification of \definitionref{defn:consistent}} \label{sect:justification-consistent}
\begin{proposition} \label{prop:justification-consistent}
    Suppose a policy profile $\bfA \in \calS_{\policy}^N$ satisfies
    \begin{enumerate}
        \item \textbf{Fairness condition.} $|\bbE[\tau_{j,k}(T)] - \bbE[\tau_{j', k}(T)]| \le o(T^\alpha)$ for all $\alpha > 0$, $j, j' \in [N]$ and $k \in [K]$.
        \item \textbf{No regret condition.} $\bbE[\regret_j(T)] \le o(T^{\alpha})$ for all $j \in [N]$ and $\alpha > 0$. 
    \end{enumerate}
    Then $\bfA$ is consistent as defined in \definitionref{defn:consistent}.
\end{proposition}

The proof is provided in \sectionref{sect:proof-consistent}.

%% file: paragraphs/appendix-experiment.tex
\section{Experimental Details} \label{sect:app-exp}
\subsection{Implementation details of different methods}
\begin{itemize}
    \item \textbf{SMAA (Ours)}. We use a hyper-parameter $\beta$ to control the strength between exploration and exploitation. Specifically, the KL index in \equationref{eq:kl-ucb} is modified as
    \begin{equation} \label{eq:kl-ucb-beta}
        \hat{b}_{j,k}(t) = \sup\left\{q \ge \hat{\mu}_{j,k}(t): \tau_{j,k}(t)\kl(\hat{\mu}_{j,k}(t), q) \le \beta \cdot f(t)\right\}.
    \end{equation}
    We search the hyper-parameter $\beta \in \{0.01, 0.02, 0.05, 0.1, 0.2, 0.5\}$.
    \item \textbf{SMAA (Ours) when $N$ and rank are unknown}. Besides the hyper-parameter $\beta$ in the exploration-exploitation phase, we introduce another hyper-parameter for this method in the Musical Chairs phase. Specifically, we set the parameter $T_0$ in \algorithmref{alg:musical-chair} as $\eta \cdot \ceil{50 K^2\log (4T)}$ and $\eta$ is searched in $\{0.01, 0.02, 0.05, 0.1, 0.2, 0.5\}$.
    \item \textbf{SelfishRobustMMAB} \citep{boursier2020selfish}. This baseline considers the setting where no player would get the reward if collisions occur. As a result, it could only apply to the scenarios $K \ge N$. It provides algorithms for each player that is robust to the deviation of a single selfish player under this setting. This method also adopts the KL-UCB to calculate the index of sub-optimal arms, as shown in \equationref{eq:kl-ucb-beta}. We also search the hyper-parameter $\beta \in \{0.01, 0.02, 0.05, 0.1, 0.2, 0.5\}$.
    \item \textbf{TotalReward} \citep{bande2021dynamic}. This baseline considers the setting where players can get the reward when collisions occur. However, the target of this baseline is to maximize the total reward of all players. When $K \ge N$, the optimal solution at each round is to make all players pull the best $N$ arms. When $K \le N$, all arms should be pulled in each round. To further reduce the regrets of each player, after assigning $k$ players to each arm, we make the rest $N - K$ players follow the Nash equilibrium in each round. The whole method is conducted in an explore-then-commit fashion. As a result, we set the algorithm to randomly explore the arms at the first $\alpha \log T$ rounds. Afterward, the algorithm commits to the optimal solution at each round. We search the hyper-parameter $\alpha \in \{100, 200, 500, 1000, 2000\}$.
    \item \textbf{ExploreThenCommit}. We further implement an explore-then-commit-based method. Specifically, we first adopt an exploration stage where all players pull the arms randomly. After the exploration stage, each player estimates the expected reward of each arm and computes the Nash equilibrium \textit{w.r.t.} the estimations. They then commit to the estimated equilibrium till the last round. The exploration phase has $\alpha \log T$ rounds where $\alpha$ is a hyper-parameter searched in $\{100, 200, 500, 1000, 2000\}$.

\end{itemize}

%% file: paragraphs/appendix-proof.tex
\section{Omitted Proofs}
\subsection{Proof of \theoremref{thrm:PNE}}
\begin{proof}
    (1) We first show that there exists a strategy profile $\bfs$ such that $\bfm(\bfs) = \bfm^*$.
    
    We only need to prove that $\sum_{k=1}^K m_k^* = N$. Note that $z^*$ exists since $h(z) \rightarrow \infty$ when $z \rightarrow 0$ and $h(z) = 0$ when $z > 1$. Moreover, under \assumptionref{assum:unique}, for any $\delta < \delta_0$ ($\delta_0$ is defined in \equationref{eq:delta-0}), $h(z) - h(z + \delta) \le 1$. As a result, there must exist $z \in \bbR^+$ such that $h(z) = N$. Therefore, $h(z^*) = N$ since $h(z)$ is right continuous. Now we can get that $\sum_{k=1}^K m^*_k = h(z^*) = N$.

    (2) We then show that any strategy profile $\bfs$ that satisfies $\bfm(\bfs) = \bfm^*$ is a Nash equilibrium.

    Suppose a player $j$ deviate from the strategy profile and change the chosen arm from $a = s_j$ to $b$ ($a \ne b$). By definition, we have $m^*_a > 0$. In addition, since $m_b^* = \floor{\mu_b / z^*}$, we have $m_b^* + 1 > \mu_b / z^*$. As a result, the reward after deviation is $\mu_b/(m_b^* + 1) < z^*$. However, the original reward is $\mu_a / n^*_a \ge z^*$, which means that the deviation only leads to the decrease in the reward of the player. Therefore, $\bfm^*$ is a Nash equilibrium of the game.
    
    We note that we have demonstrated the existence of the Nash equilibrium through points (1) and (2).
    
    (3) Afterward, we show that for any strategy profile $\bfs$ that is a Nash equilibrium, $\bfm(\bfs) = \bfm^*$.
    
    Under \assumptionref{assum:unique}, suppose there exists a Nash equilibrium $\bfs$ such that $\bfm(\bfs) \ne \bfm^*$. Let $\bfm' = \bfm(\bfs)$. Then there exist two arms $a$ and $b$ such that $\bfm^*_a > \bfm'_a$ and $\bfm^*_b < \bfm'_b$. Using similar techniques with the first part, it is easy to show that a player deviate from arm $b$ to arm $a$ in the profile $\bfs$ will lead to an increase in the reward, which results in a contradiction.
    
    Now the claim follows.
\end{proof}

\subsection{Proofs of \propositionref{prop:poa}} \label{sect:proof-poa}
\begin{proof}
    (1) Lower bound $1$. The PoA is naturally lower bounded by $1$ and can take on this value when the expected rewards of the top $\min(N, K)$ arms are very close, which is possible for any values of $N$ and $K$.

    (2) Upper bound $(N + \min\{K,N\} - 1) / N$. Due to the definition of $z^*$ in Theorem 3.1, we have $W^{\text{PNE}} / N \ge z^* > \mu_k$ for any $k \not\in \mathcal{M}^*$. As a result, for any arm $k \in \calC^* \backslash \mathcal{M}^*$, $\mu_k < W^{\text{PNE}} / N$. In addition, due to the property of Nash equilibrium demonstrated in Theorem 3.1, all arms in the equilibrium have the highest expected rewards. As a result, $\mathcal{M}^* \subseteq \calC^*$. Hence, 
    \begin{equation}
        W^{\text{MAX}} = W^{\text{PNE}} + \sum_{k \in \calC^* \backslash \mathcal{M}^*} \mu_k \le W^{\text{PNE}} + |\calC^* \backslash \mathcal{M}^*| W^{\text{PNE}} / N \le W^{\text{PNE}} \left(1 + (\min(N, K) - 1) / N\right).     
    \end{equation}
    As a result,
    \begin{equation}
        \text{PoA} = W^{\text{MAX}} / W^{\text{PNE}}  \le (N + \min\{K,N\} - 1) / N.
    \end{equation}
    In addition, the exact value of PoA can be taken any close to $(N + \min\{K,N\} - 1) / N$ by setting $\mu_1 = 1$ and the rewards of other arms very close to but smaller than $1 / N$.
\end{proof}

\subsection{Proofs in \sectionref{sect:theory}} \label{sect:proof-theory}
\subsubsection{Basic Notations and Lemmas}
Fix $\delta$ such that $0 < \delta < \delta_0 / 2$. Let $K' = N\ceil{K/N}$. Define the following random sets
\begin{equation} \label{eq:time-sets}
    \begin{aligned}
        \calA_j & = \left\{t > K': \calM^* \backslash \tilde{\calM}_j(t) \ne \emptyset \right\}, \\
        \calB_j & = \left\{t > K': \exists k \in \tilde{\calM}_j(t), \left|\tilde{\mu}_{j,k}(t) - \mu_k\right| \ge \delta\right\}, \\
        \calC_j & = \left\{t > K': \exists k \in \calM^*, \tilde{b}_{j,k}(t) < \mu_k \right\}, \\
        \calD_j & = \left\{t \in \calA_j \backslash (\calB_j \cup \calC_j) : \exists k \in \calM^* \backslash \tilde{\calM}_j(t), \left|\tilde{\mu}_{j,k}(t) - \mu_k\right| \ge \delta\right\}, \\
        \calJ & = \left\{t > K': \exists j \in [K], t \in (\calA_j \cup \calB_j)\right\}.
    \end{aligned}
\end{equation}

Let $l^*$ be the optimal list of the arm chosen in a block \textit{w.r.t.} the expected rewards $\{\mu_k\}_{k=1}^K$, using the similar procedures with that in \equationref{eq:sequence-in-block}. Formally, we sort the arms in the Nash equilibrium $\calM^*$ according to $r^*_k = \mu^*_k / m^*_k$ by descending order and get the indices of the arms $k^*_1, k^*_2, \dots, k^*_{|\calM^*|}$. Then we sequentially choose these arms and each arm $k^*_i$ are further chosen by $m^*_i$ times. The arms list $l^*$ is given by
\begin{equation} \label{eq:sequence-in-block-optimal}
    l^* = \left(\underbrace{k^*_1, \dots, k^*_1}_{m^*_{k^*_1} \text{ times}}, \dots, \underbrace{k^*_{|\calM^*|}, \cdots, k^*_{|\calM^*|}}_{m^*_{k^*_{|\calM^*|}} \text{ times}}\right).
\end{equation}

\begin{lemma} \label{lemma:z-star}
    Under \assumptionref{assum:unique}, we have
    \begin{equation}
        z^* = \min_{k \in [K], m^*_k > 0} \mu_k/m^*_k,
    \end{equation}
    where $z^*$ is given by \equationref{eq:z-star}.
\end{lemma}
\begin{lemma} \label{lemma:clean-event}
    Under \assumptionref{assum:unique}, suppose $t > K'$ and $t \not\in (\calA_j \cup \calB_j)$. Let $k_0$ be the last element in $\tilde{l}_j(t)$. Then $l^* = \tilde{l}_j(t)$, and $|\tilde{r}_{j,k_0}(t) - z^*| \le \delta$ where $z^*$ is given by \equationref{eq:z-star}.
\end{lemma}
\begin{remark}
    The proof is provided in \appendixref{sect:proof-lemma-clean-event}. \lemmaref{lemma:clean-event} demonstrates that when $t \not\in (\calA_j \cup \calB_j)$, player $j$ could find the Nash equilibrium of the game as shown in \theoremref{thrm:PNE} and choose arms according to $l^*$ in \equationref{eq:sequence-in-block-optimal}.
\end{remark}

\begin{lemma} \label{lemma:subseteq}
    Under \assumptionref{assum:unique}, we have $\calA_j \cup \calB_j \subseteq \calB_j \cup \calC_j \cup \calD_j$.
\end{lemma}

\begin{lemma} \label{lemma:subset-expectation}
    Under \assumptionref{assum:unique}, we have $\bbE[|\calB_j \cup \calC_j \cup \calD_j|] \le 8N^2K(12K + \delta^{-2})$.
\end{lemma}

\begin{lemma} \label{lemma:bad-arms}
    Under \assumptionref{assum:unique}, define
    \begin{equation} \label{eq:calG}
        \calG_{j,k} \triangleq \left\{K' < t \le T: t \not\in (\calA_j \cup \calB_j), \pi_j(t) = k\right\}
    \end{equation}
    for any $k \not\in \calM^*$. Then
    \begin{equation}
        \bbE[|\calG_{j,k}|] \le \frac{\log T + 4 \log (\log T)}{\kl(\mu_k + \delta, z^* - \delta)} + 5 + 2\delta^{-2}.
    \end{equation}
\end{lemma}

\begin{remark}
    The proofs of \lemmaref{lemma:subseteq}, \lemmaref{lemma:subset-expectation}, and \lemmaref{lemma:bad-arms} follow the techniques in \citep{wang2020optimal} and details are provided in \appendixref{sect:proof-lemma-subseteq}, \appendixref{sect:proof-lemma-subset-expectation}, and \appendixref{sect:proof-lemma-bad-arms}, respectively. We highlight that we consider a more complex setting since players need to calculate the Nash equilibrium in each round according to \theoremref{thrm:PNE}. As a result, the details of the proofs become more complicated.
\end{remark}

\subsubsection{Proof of \theoremref{thrm:main-regret}} \label{sect:proof-main-regret}
\begin{proof}
    Consider the set $\calJ$ defined in \equationref{eq:time-sets}, according to \lemmaref{lemma:subseteq} and \lemmaref{lemma:subset-expectation}, we have
    \begin{equation}
        \bbE[|\calJ|] \le \sum_{j=1}^N \bbE[|\calA_j \cup \calB_j|] \le \sum_{j=1}^N \bbE[|\calB_j \cup \calC_j \cup \calD_j|] \le 8N^3K(12K + \delta^{-2}).
    \end{equation}
    As a result, we have
    \begin{equation}
        \begin{aligned}
            \bbE[\regret_j(T)] = & \, \bbE\left[\sum_{t=1}^T\left(\max_{k \in [k]}\frac{\mu_{k}}{M_{k}(t) + \bbI[\pi_j(t) \ne k]} - R_j(t)\right)\right] \\
            \le & \, K' + \bbE\left[\sum_{t=K'+1}^T\bbI[t \in \calJ]\left(\max_{k \in [k]}\frac{\mu_{k}}{M_{k}(t) + \bbI[\pi_j(t) \ne k]} - R_j(t)\right)\right] \\
            & \, + \bbE\left[\sum_{t=K'+1}^T\bbI[t \not\in \calJ]\left(\max_{k \in [k]}\frac{\mu_{k}}{M_{k}(t) + \bbI[\pi_j(t) \ne k]} - R_j(t)\right)\right] \\
            \le & \, K' + \bbE[|\calJ|] + \bbE\left[\sum_{t=K'+1}^T\bbI[t \not\in \calJ]\left(\max_{k \in [k]}\frac{\mu_{k}}{M_{k}(t) + \bbI[\pi_j(t) \ne k]} - R_j(t)\right)\right] \\
            \le & \, 8N^3K(13K + \delta^{-2}) + \bbE\left[\sum_{t=K'+1}^T\bbI[t \not\in \calJ]\left(\max_{k \in [k]}\frac{\mu_{k}}{M_{k}(t) + \bbI[\pi_j(t) \ne k]} - \mu_{\pi_j(t)}\right)\right].
        \end{aligned}
    \end{equation}
    When $t > K'$ and $t \not\in J$, according to \lemmaref{lemma:clean-event}, each player $j$ could find the correct Nash equilibrium $\calM^*$ and calculate the correct arms list $l^*$. Let $k_0$ be the last element in $l^*$. As a result, at most one player do not follow the equilibrium $\calM^*$ by the construction of \algorithmref{alg:SMAA}. Formally, this indicates that $\forall k \in \calM^* \backslash \{k_0\}$, $M_{k}(t) = m_k^*$ and $M_{k_0}(t) \ge m_{k_0}^* - 1$. As a result, when $\pi_j(t) \in \calM^*$, it will not increase the reward if deviation by the property of the equlibrium and the choice of $k_0$, \textit{i.e.},
    \begin{equation}
        \bbI[t \not\in \calJ, \pi_j(t) \in \calM^*]\left(\max_{k \in [k]}\frac{\mu_{k}}{M_{k}(t) + \bbI[\pi_j(t) \ne k]} - \mu_{\pi_j(t)}]\right) = 0.
    \end{equation}
    In addition, when $\pi_j(t) \not\in \calM^*$, because $\forall k \in \calM^* \backslash \{k_0\}$, $M_{k}(t) = m_k^*$ and $M_{k_0}(t) \ge m_{k_0}^* - 1$, it will maximize the reward if deviating to arm $k_0$. As a result, $\max_{k \in [k]}\mu_{k}/(M_{k}(t) + \bbI[\pi_j(t) \ne k]) = z^*$ in this case. Therefore, according to \lemmaref{lemma:bad-arms} and the definition of $\calG_{j,k}$ in \equationref{eq:calG}, we have
    \begin{equation} \label{eq:proof-main-regret-last}
        \begin{aligned}
            \bbE[\regret_j(T)] & \le 8N^3K(13K + \delta^{-2}) +  \bbE\left[\sum_{k\not\in \calM^*}\sum_{t=K'+1}^T\bbI[t \not\in \calJ, t \in \calG_{j,k}]\left(\max_{k \in [k]}\frac{\mu_{k}}{M_{k}(t) + \bbI[\pi_j(t) \ne k]} - \mu_{\pi_j(t)}\right)\right] \\
            & = 8N^3K(13K + \delta^{-2}) + \bbE\left[\sum_{k\not\in \calM^*}\sum_{t=K'+1}^T\bbI[t \not\in \calJ, t \in \calG_{j,k}]\left(z^* - \mu_{k}\right)\right] \\
            & \le 8N^3K(13K + \delta^{-2}) + \bbE\left[\sum_{k\not\in \calM^*}\left(z^* - \mu_{k}\right)\sum_{t=K'+1}^T\bbI[t \in \calG_{j,k}]\right] \\
            & \le 8N^3K(13K + \delta^{-2}) + \sum_{k \not\in\calM^*}\frac{(z^*-\mu_k)(\log T + 4\log (\log T))}{\kl(\mu_k+\delta, z^*-\delta)} + (5 + 2\delta^{-2})K \\
            & \le 10N^3K(13K + \delta^{-2}) + \sum_{k \not\in\calM^*}\frac{(z^*-\mu_k)(\log T + 4\log (\log T))}{\kl(\mu_k+\delta, z^*-\delta)}.
        \end{aligned}
    \end{equation}
    Now the claim follows.
\end{proof}

\subsubsection{Proof of \theoremref{thrm:main-noneq}} \label{sect:proof-main-noneq}
\begin{proof}
    Let $k_0$ be the last element in $l^*$. Similar to the proof of \theoremref{thrm:main-regret} in \appendixref{sect:proof-main-regret}, we have $\bbE[\calJ] \le 8N^3K(12K + \delta^{-2})$. In addition, when $t > K'$ and $t \not\in J$, we have $\forall k \in \calM^* \backslash \{k_0\}$, $M_{k}(t) = m_k^*$ and $M_{k_0}(t) \ge m_{k_0}^* - 1$. As a result, if players do not achieve the Nash equilibrium at round $t$, there must exist a player $j$ such that $\pi_j(t) \not\in \calM^*$. As a result, according to \lemmaref{lemma:bad-arms} and the definition of $\calG_{j,k}$ in \equationref{eq:calG}, we have
    \begin{equation}
        \begin{aligned}
            \bbE[\noneq(T)] & = \bbE\left[\sum_{t=1}^T \bbI\left[\exists k \in [K], M_k(t) \ne m^*_k\right]\right] \\
            & \le K' + \bbE\left[\sum_{t=K'+1}^T \bbI\left[\exists k \in [K], M_k(t) \ne m^*_k, t \in \calJ\right]\right] + \bbE\left[\sum_{t=K'}^T \bbI\left[\exists k \in [K], M_k(t) \ne m^*_k, t \not\in \calJ\right]\right] \\
            & \le K' + \bbE[|\calJ|] + \bbE\left[\sum_{t=K'}^T \bbI\left[\exists j \in [N], \pi_j(t) \not\in \calM^*, t \not\in \calJ\right]\right] \\
            & \le 8N^3K(13K + \delta^{-2}) + \sum_{j=1}^N\sum_{k\not\in \calM^*}\bbE\left[\sum_{t=K'}^T\bbI[t \not\in \calJ, \pi_j(t) = k]\right] \\
            & \le 8N^3K(13K + \delta^{-2}) + \sum_{j=1}^N\sum_{k \not\in \calM^*} \bbE[|\calG_{j,k}|] \\
            & \le 8N^3K(13K + \delta^{-2}) + N\sum_{k \not\in \calM^*}\frac{\log T + 4\log(\log T)}{\kl(\mu_k + \delta, z^* - \delta)} + (5 + 2\delta^{-2})NK \\
            & \le 10N^3K(13K + \delta^{-2}) + N\sum_{k \not\in \calM^*}\frac{\log T + 4\log(\log T)}{\kl(\mu_k + \delta, z^* - \delta)}.
        \end{aligned}
    \end{equation}
    Now the claim follows.
\end{proof}

\subsubsection{Proof of \theoremref{thrm:selfish}} 
\label{sect:proof-selfish}
\begin{proof}
    For the $\epsilon$-Nash equilibrium part, we note that \algorithmref{alg:SMAA} only uses the information of $X_{\pi_j(t)}(t)$, which means that the behaviors of strategic players will not affect the policy of any other player. As a result, the reward of a strategic player deviate from \algorithmref{alg:SMAA} to other algorithm is at most the regret demonstrated in \theoremref{thrm:main-regret}. Hence, the profile where all players follow \algorithmref{alg:SMAA} is an $\epsilon$-Nash equilibrium with
    \begin{equation}
        \epsilon = \sum_{k \not\in \calM^*} \frac{(z^* - \mu_k)(\log T + 4 \log (\log T))}{\kl(\mu_k + \delta, z^* - \delta)} + 10N^3K(13K + \delta^{-2}).
    \end{equation}
    
    Consider the $(\beta, \epsilon + \beta\gamma)$-stable part. Suppose player $j$ is strategic and player $i$ follows \algorithmref{alg:SMAA}. When $t \in \calJ$ or $t \le K'$, the increase of player $j$'s reward is at most $1$ and the decrease in player $i$'s reward is at most 1. Let $k_0$ be the last element in $l^*$.
    
    We first show that when $t > K'$ and $t \not\in \calJ$, the loss of player $i$'s reward in each round due to the deviation of player $j$ is at most $z^*$. When $t > K'$ and $t \not\in \calJ$, according to the proof of \theoremref{thrm:main-regret} in  \appendixref{sect:proof-main-regret}, we have $\forall k \in \calM^* \backslash \{k_0\}$, $M_{k}(t) = m_k^*$ and $M_{k_0}(t) \ge m_{k_0}^* - 1$. We consider different cases of the values of $\pi_i(t)$
    \begin{itemize}
        \item Consider the case when $\pi_i(t) \in \calM^* \backslash \{k_0\}$. Then if $\pi_i(t) = \pi_j(t)$, player $j$ can not lead to a decrease in player $i$' reward. If $\pi_i(t) \ne \pi_j(t)$, player $j$ will deviate to pull arm $\pi_i(t)$ to make player $i$ worse. In this case, when $\pi_i(t) \ne k_0$, the loss of player $i$'s reward is
        \begin{equation}
            \frac{\mu_k}{m_k^*} - \frac{\mu_k}{m_k^* + 1} = \frac{\mu_k}{m_k^*(m_k^*+1)} \le \frac{\mu_k}{m_k^*+1} < z^*.
        \end{equation}
        \item Consider the case when $\pi_i(t) = k_0$. If $M_{k_0}(t) = m^*_{k_0} - 1$, then $m^*_{k_0} > 1$ and the loss of player $i$'s reward is
        \begin{equation}
            \frac{\mu_{k_0}}{m^*_{k_0} - 1} - \frac{\mu_{k_0}}{m^*_{k_0}} = \frac{\mu_{k_0}}{m^*_{k_0}(m^*_{k_0} - 1)} \le \frac{\mu_{k_0}}{m^*_{k_0}} = z^*.
        \end{equation}
        \item Consider the case when $\pi_i(t) \not\in \calM^*$. Then the loss of player $i$'s reward is $\mu_{\pi_i(t)} / 2 < z^*$.
    \end{itemize}
    
    We then consider the change of player $j$'s reward by deviation when $t > K'$ and $t \not\in \calJ$. When $\pi_j(t) \not\in \calM^*$, player $j$ could increase the reward by at most $z^* - \mu_{\pi_j(t)}$. When $\pi_j(t) \in \calM^*$, player $j$ will also incur a loss at least $\delta_0$.
    
    Suppose the player $j$ deviate to policy $s'$, let $\calQ$ be the set of rounds that player $j$ choose to deviate from \algorithmref{alg:SMAA}. Let
    \begin{equation}
        \begin{aligned}
            & \calQ_1 = \left\{t \in \calQ: t \in \calJ\cup\{1, 2, \dots, K'\} \right\}, \calQ_2 = \{t \in \calQ: t \not \in \calJ, t > K', \pi_j(t) \not\in \calM^*\}, \\
            & \calQ_3 = \{t \in \calQ: t \not \in \calJ, t > K', \pi_j(t) \in \calM^*\}.
        \end{aligned}
    \end{equation}
    Then $\calQ_1$, $\calQ_2$, and $\calQ_3$ are disjoint and $\calQ = \calQ_1 \cup \calQ_2 \cup \calQ_3$.
    Then
    \begin{equation}
        \begin{aligned}
            \reward_i(T; s', \bfs_{-j}) & \ge \reward_i(T; \bfs) - |\calQ_1| - z^*(|\calQ_2| + |\calQ_3|), \\
            \reward_j(T; s', \bfs_{-j}) & \le \reward_j(T; \bfs) + |\calQ_1| + \left(\sum_{k \not \in \calM^*}(z^* - \mu_k)|\{t \in \calQ_2: \pi_j(t) = k\}|\right) - \delta_0 |\calQ_3|.
        \end{aligned}
    \end{equation}
    Taking expectations, according to \lemmaref{lemma:bad-arms} and the definition of $\calG_{j,k}$ in \equationref{eq:calG}, we have
    \begin{equation}
        \begin{aligned}
             & \, \bbE[\reward_i(T; s', \bfs_{-j})] - \bbE[\reward_i(T; \bfs)] \\
             \ge & \, -\bbE[|\calQ_1|] - z^*\bbE[|\calQ_2|] - z^* \bbE[|\calQ_3|] \\
             \ge & \, -8N^3K(13K + \delta^{-2}) - z^* \sum_{k \not\in \calM^*}\bbE[|\calG_{j,k}|] - z^* \bbE[|\calQ_3|] \\
             \ge & \, -8N^3K(13K + \delta^{-2}) - z^*\left(K(5 + 2\delta^{-2}) + \sum_{k \not\in \calM^*}\frac{\log T + \log(\log T)}{\kl(\mu_k + \delta, z^* - \delta)}\right) - z^*\bbE[|\calQ_3|] \\
             \ge & \, -10N^3K(13K + \delta^{-2}) - \sum_{k \not\in \calM^*}\frac{\log T + \log(\log T)}{\kl(\mu_k + \delta, z^* - \delta)} - z^*\bbE[|\calQ_3|]
        \end{aligned}
    \end{equation}
    In addition,
    \begin{equation}
        \begin{aligned}
            & \, \bbE[\reward_j(T; s', \bfs_{-j})] - \bbE[\reward_j(T; \bfs)] \\
            \le & \, \bbE[|\calQ_1|] + \left(\sum_{k \not \in \calM^*}(z^* - \mu_k)\bbE[|\{t \in \calQ_2: \pi_j(t) = k\}|]\right) - \delta_0 \bbE[|\calQ_3|] \\
            \le & \, \bbE[|\calQ_1|] + \left(\sum_{k \not \in \calM^*}(z^* - \mu_k)\bbE[|\calG_{j,k}|]\right) - \delta_0 \bbE[|\calQ_3|] \\
            \le & \, 10N^3K(13K + \delta^{-2}) + \sum_{k \not\in\calM^*}\frac{(z^*-\mu_k)(\log T + 4\log (\log T))}{\kl(\mu_k+\delta, z^*-\delta)} - \delta_0 \bbE[|\calQ_3|].
        \end{aligned}
    \end{equation}
    Here the last inequality is due to \equationref{eq:proof-main-regret-last}. Because
    \begin{equation}
            \gamma = 10N^3K(13K + \delta^{-2}) + \sum_{k \not\in \calM^*}\frac{\log T + \log(\log T)}{\kl(\mu_k + \delta, z^* - \delta)},
    \end{equation}
    we have
    \begin{equation}
        \bbE[\reward_i(T; s', \bfs_{-j})] - \bbE[\reward_i(T; \bfs)] \ge -\gamma - z^*\bbE[|\calQ_3|] \quad \text{and} \quad \bbE[\reward_j(T; s', \bfs_{-j})] - \bbE[\reward_j(T; \bfs)] \le \epsilon - \delta_0 \bbE[|\calQ_3|].
    \end{equation}
    As a result, for any $u \in \bbR^+$, if $\bbE[\reward_i(T; s', \bfs_{-j})] - \bbE[\reward_i(T; \bfs)] \le -u$, we have $\bbE[|\calQ_3|] \ge (u - \gamma) / z^*$. Then
    \begin{equation}
        \bbE[\reward_j(T; s', \bfs_{-j})] - \bbE[\reward_j(T; \bfs)] \le \epsilon - \delta_0 \frac{u - \gamma}{z^*} = \epsilon + \frac{\delta_0\gamma}{z^*} - \frac{\delta_0}{z^*}u.
    \end{equation}
    Now the claim follows.
\end{proof}

\subsubsection{Proof of \theoremref{thrm:lower-bound}} \label{sect:proof-lower-bound}
\begin{proof}
    The proof follows the techniques of the proof of Lemma 3 in \citep{liu2010distributed} and Theorem 1 in \citep{anantharam1987asymptotically}.
    
    Let any problem instance $\bfxi = (\xi_1, \xi_2, \dots, \xi_K) \in \calI_{N,K}^{\ber}$. Here $\xi_1, \xi_2, \dots, \xi_K$ are Bernoulli distributions with expectations $\mu_1, \mu_2, \dots, \mu_K$. $z^*$ and $\bfm^*$ are calculated according to \theoremref{thrm:PNE}. $\calM^* = \{k \in [K]: m^*_k > 0\}$. $\delta_0$ is given by \equationref{eq:delta-0}. Consider any sub-optimal arm $k$ such that $\mu_k < z^*$. As a result, $m^*_k = 0$.

    For any $\beta \in (0,1)$, choose a parameter $\lambda$ such that
    \begin{equation} \label{eq:proof-thrm-lower-bound-0}
        z^* < \lambda < z^* + \delta_0, \quad \left|\kl(\mu_k, \lambda) - \kl(\mu_k, z^*)\right| \le \beta \kl(\mu_k, z^*), \quad \text{and} \quad \bfxi' \triangleq (\xi_1, \xi_2, \dots, \xi_{k-1}, \xi_k', \xi_{k+1}, \dots, \xi_K) \in \Xi.
    \end{equation}
    Here $\xi_k'$ is the Bernoulli distribution with expectation $\lambda$. We use $z'^*$, $\bfm'^*$, and $\calM'^*$ to denote the Nash equilibrium \textit{w.r.t.} the problem instance $\bfxi'$. We use $\bbP_{\bfxi}[E]$ and $\bbP_{\bfxi'}[E]$ to denote the probability of an event $E$ when the problem instance is $\bfxi$ and $\bfxi'$, respectively.
    
    According to \theoremref{thrm:PNE} and \lemmaref{lemma:z-star}, let $k_0$ be the arm in $\calM^*$ such that $\mu_{k_0} / m^*_{k_0} = z^*$. Since $z^* < \lambda < z^* + \delta$, we have that
    \begin{equation}
        \forall k' \in \calM^* \backslash \{k_0\}, \quad m^*_{k'} \le \frac{\mu_{k'}}{\floor{\frac{\mu_{k'}}{m^*_{k'}}}} < \frac{\mu_{k'}}{z^* + \delta} < \frac{\mu_{k'}}{z^* + \delta} < \frac{\mu_{k'}}{\lambda} < \frac{\mu_{k'}}{z^*} < m^*_{k'} + 1.
    \end{equation}
    As a result, $\floor{\mu_{k'} / \lambda} = m^*_{k'}$. In addition, $\floor{\mu_{k_0} / \lambda} = m^*_{k_0} - 1$. As a result,
    \begin{equation}
        \sum_{k'=1}^K\floor{\frac{\mu'_{k'}}{\lambda}} = \left(\sum_{k' \in \calM^*} \floor{\frac{\mu'_{k'}}{\lambda}}\right) + \floor{\frac{\lambda}{\lambda}} + \left(\sum_{k' \not\in \calM^* \cup \{k\}} \floor{\frac{\mu'_{k'}}{\lambda}}\right) = N - 1 + 1 + 0 = N.
    \end{equation}
    and we have $m'^*_k = 1$. Therefore, since algorithm $A$ is consistent, we have $T / N - \bbE_{\bfxi'}[\tau_{j,k}(T)] \le o(T^\alpha)$ for any $\alpha \in (0, \beta)$. Define a random function $c(T)$ as
    \begin{equation}
        c(T) \triangleq \max\left\{\bbE_{\bfxi'}\left[\frac{T}{N} - \tau_{j,k}(T)\right], 0\right\} - \frac{T}{N} + \tau_{j,k}(T).
    \end{equation}
    We have $\bbE_{\bfxi'}[c(T)] \ge 0$ and
    \begin{equation} \label{eq:proof-thrm-lower-bound-1}
        \bbE_{\bfxi'}\left[c(T) + \frac{T}{N} - \tau_{j,k}(T)\right] = o(T^\alpha).
    \end{equation}
    Because $c(T) + T / N - \tau_{j,k}(T) \ge 0$ a.e., we have
    \begin{equation} \label{eq:proof-thrm-lower-bound-2}
        \begin{aligned}
            \bbE_{\bfxi'}\left[c(T) + \frac{T}{N} - \tau_{j,k}(T)\right] = & \, \bbE_{\bfxi'}\left[c(T) + \frac{T}{N} - \tau_{j,k}(T) \Big| \tau_{j,k}(T) < \frac{(1 - \beta)\log T}{\kl(\mu_k, \lambda)}\right]\bbP_{\bfxi'}\left[\tau_{j,k}(T) < \frac{(1 - \beta)\log T}{\kl(\mu_k, \lambda)}\right] \\
            & \, + \bbE_{\bfxi'}\left[c(T) + \frac{T}{N} - \tau_{j,k}(T) \Big| \tau_{j,k}(T) \ge \frac{(1 - \beta)\log T}{\kl(\mu_k, \lambda)}\right]\bbP_{\bfxi'}\left[\tau_{j,k}(T) \ge \frac{(1 - \beta)\log T}{\kl(\mu_k, \lambda)}\right] \\
            \ge & \, \bbE_{\bfxi'}\left[c(T) + \frac{T}{N} - \tau_{j,k}(T) \Big| \tau_{j,k}(T) < \frac{(1 - \beta)\log T}{\kl(\mu_k, \lambda)}\right]\bbP_{\bfxi'}\left[\tau_{j,k}(T) < \frac{(1 - \beta)\log T}{\kl(\mu_k, \lambda)}\right] \\
            \ge & \, \left(\bbE_{\bfxi'}[c(T)] + \frac{T}{N} - \frac{(1 - \beta)\log T}{\kl(\mu_k, \lambda)}\right)\bbP_{\bfxi'}\left[\tau_{j,k}(T) < \frac{(1 - \beta)\log T}{\kl(\mu_k, \lambda)}\right].
        \end{aligned}
    \end{equation}
    
    Let $S_{k, 1}, S_{k, 2}, \dots$ be the independent observations from arm $k$. Define
    \begin{equation}
        L_t \triangleq \sum_{i=1}^t \log\left(\frac{\bbP_{\bfxi}[X_k = S_{k,i}]}{\bbP_{\bfxi'}[X_k = S_{k,i}]}\right)
    \end{equation}
    and event
    \begin{equation}
        C \triangleq \left\{\tau_{j,k}(T) < \frac{(1 - \beta)\log T}{\kl(\mu_k, \lambda)}, L_{\tau_{j,k}(T)} \le \left(1-\frac{\alpha + \beta}{2}\right)\log T\right\}.
    \end{equation}
    Combining \equationsref{eq:proof-thrm-lower-bound-1} and \eqref{eq:proof-thrm-lower-bound-2}, we have
    \begin{equation}
        \begin{aligned}
            \bbP_{\bfxi'}[C] & \le \bbP_{\bfxi'}\left[\tau_{j,k}(T) < \frac{(1 - \beta)\log T}{\kl(\mu_k, \lambda)}\right] \le \frac{\bbE_{\bfxi'}\left[c(T) + T/N - \tau_{j,k}(T)\right]}{\bbE_{\bfxi'}[c(T)] + T/N - (1 - \beta)\log T/\kl(\mu_k, \lambda)} \\
            & \le \frac{o(T^\alpha)}{T/N - (1 - \beta)\log T/\kl(\mu_k, \lambda)} = o(T^{\alpha-1}).
        \end{aligned}
    \end{equation}
    Let $C_s \triangleq \{\tau_{j,k}(T) = s, L_s \le (1 - (\alpha + \beta) / 2)\log T\}$. We have
    \begin{equation}
        \begin{aligned}
            \bbP_{\bfxi'}[C_s] & = \int_{\{\tau_{j,k}(T) = s, L_s \le (1 - (\alpha + \beta) / 2)\log T\}} \mathrm{d} \bbP_{\bfxi'} = \int_{\{\tau_{j,k}(T) = s, L_s \le (1 - (\alpha + \beta) / 2)\log T\}} \prod_{i=1}^s \frac{\bbP_{\bfxi'}[X_k = S_{k,i}]}{\bbP_{\bfxi}[X_k = S_{k,i}]}\mathrm{d} \bbP_{\bfxi} \\
            & \ge \int_{\{\tau_{j,k}(T) = s, L_s \le (1 - (\alpha + \beta) / 2)\log T\}}\exp\left(-(1-(\alpha + \beta) / 2)\log T\right)\mathrm{d}\bbP_{\bfxi} = T^{(\alpha + \beta) / 2 - 1}\bbP_{\bfxi}[C_s].
        \end{aligned}
    \end{equation}
    Since $C_s$ $(1 \le s < (1 - \beta)\log T / \kl(\mu_k, \lambda))$ are disjoint, we have
    \begin{equation} \label{eq:proof-thrm-lower-bound-3}
        \bbP_{\bfxi}[C] \le T^{1 - (\alpha + \beta) / 2} \bbP_{\bfxi'}[C] = o\left(T^{\frac{\alpha - \beta}{2}}\right) \rightarrow 0, \quad \text{when } T \rightarrow \infty.
    \end{equation}
    
    In addition, by the law of large numbers, we have that under $\bbP_{\bfxi}$, when $t \rightarrow \infty$, $L_t / t \rightarrow \kl(\mu_k, \lambda) > 0$. As a result, under $\bbP_{\bfxi}$, when $t \rightarrow \infty$, $\max_{i \le t} L_i / t \rightarrow \kl(\mu_k, \lambda) > 0$. Therefore, since $\beta > \alpha$, we have $(1 - (\alpha + \beta) / 2) > (1 - \beta)$ and
    \begin{equation}
        \lim_{T \rightarrow \infty} \bbP_{\bfxi}\left[L_i > (1 - (\alpha + \beta) / 2) \log T \text{ for some } i < \frac{(1 - \beta)\log T}{\kl(\mu_k, \lambda)}\right] = 0.
    \end{equation}
    As a result,
    \begin{equation} \label{eq:proof-thrm-lower-bound-4}
        \lim_{T \rightarrow \infty} \bbP_{\bfxi}\left[\tau_{j,k}(T) < \frac{(1 - \beta)\log T}{\kl(\mu_k, \lambda)}, L_{\tau_{j,k}(T)} > \left(1 - \frac{\alpha + \beta}{2}\right) \log T\right] = 0.
    \end{equation}
    Combining \equationsref{eq:proof-thrm-lower-bound-3} and \eqref{eq:proof-thrm-lower-bound-4}, we have
    \begin{equation}
        \lim_{T \rightarrow \infty}\bbP_{\bfxi}\left[\tau_{j,k}(T) < \frac{(1 - \beta)\log T}{\kl(\mu_k, \lambda)}\right] = 0
    \end{equation}
    According to \equationref{eq:proof-thrm-lower-bound-0}, we have $\kl(\mu_k, \lambda) \ge (1 - \beta)\kl(\mu_k, z^*)$. Therefore,
    \begin{equation}
        \lim_{T\rightarrow \infty}\bbP_{\bfxi}\left[\tau_{j,k}(T) < \frac{(1 - \beta) \log T}{(1 + \beta) \kl(\mu_k, z^*)}\right] = 0.
    \end{equation}
    Now the claim for $\tau_{j,k}(T)$ follows by letting $\beta \rightarrow 0$.
    
    For the regret part, since for each arm $k$ that is not in the Nash equilibrium $\calM^*$, the regret for deviation is at least $z^* - \mu_k$. Therefore, we have
    \begin{equation}
        \begin{aligned}
            \regret_j(T) & = \bbE\left[\sum_{t=1}^T\max_{k \in [k]}\frac{\mu_{k}}{M_{k}(t) + \bbI[\pi_j(t) \ne k]} - \sum_{t=1}^TR_j(t)\right] \\
            & \ge \sum_{k \not\in \calM^*} \bbE\left[\sum_{t \in [T], \pi_{j}(t) = k}\left(\max_{k \in [k]}\frac{\mu_{k}}{M_{k}(t) + \bbI[\pi_j(t) \ne k]} - R_j(t)\right)\right] \\
            & \ge \sum_{k \not\in \calM^*}\bbE[\tau_{j,k}(T)] (z^* - \mu_k).
        \end{aligned}
    \end{equation}
    Now the claim for $\regret_j(T)$ follows by applying \equationref{eq:tau-lower-bound}.
\end{proof}

\subsubsection{Proof of \lemmaref{lemma:z-star}}
\begin{proof}
    On the one hand, for any $k$ such that $m^*_k > 0$, since $m^*_k = \floor{\mu_k / z^*}$, we have $m^*_k \le \mu_k / z^*$ and $z^* \le \mu_k / m^*_k$. As a result, $z^* \le \min_{k \in [K], m^*_k > 0} \mu_k/m^*_k$. On the other hand, let $z = \min_{k \in [K], m^*_k > 0} \mu_k/m^*_k$. For any for any $k$ such that $m^*_k > 0$, we have that $\mu_k / m^*_k \ge z$. Hence, we have $\mu_k / z \ge m^*_k$. As a result,
    \begin{equation}
        h(z) = \sum_{k = 1}^K \floor{\frac{\mu_k}{z}} \ge \sum_{k=1}^K m^*_k = N.
    \end{equation}
    By the definition of $z^*$, we have $z^* \ge z$. To conclude, we have $z^* \ge \min_{k \in [K], m^*_k > 0} \mu_k/m^*_k$ and $z^* \le \min_{k \in [K], m^*_k > 0} \mu_k/m^*_k$.
\end{proof}

\subsubsection{Proof of \lemmaref{lemma:clean-event}} \label{sect:proof-lemma-clean-event}
\begin{proof}
    Let
    \begin{equation}
        \tilde{z}_j(t) = \sup \left\{z > 0 : \tilde{h}_{j,t}(z) \triangleq \sum_{k=1}^K \floor{\frac{\tilde{\mu}_{j,k}(t)}{z}} \ge N\right\}
    \end{equation}
    and we have $\tilde{z}_j(t) = \tilde{r}_{j,k}(t)$ according to \theoremref{thrm:PNE} and \lemmaref{lemma:z-star}.
    
    (1) We first show that $\tilde{l}_j(t) = l^*$.
    
    Since $t \not\in \calA_j$, we have $\calM^* \subseteq \tilde{\calM}_j(t)$. Let $z_0 = z^* - \delta$. On the one hand, Fix any $k \in \tilde{\calM}_j(t)$. Under \assumptionref{assum:unique}, according to \theoremref{thrm:PNE}, we know that $z^* > \mu_k / (m^*_k+1)$ and there exists $k' \in \calM^*$ such that $z^* = \mu_{k'} / m^*_{k'}$. By the choice of $\delta$, we have
    \begin{equation}
        z^* > \frac{\mu_k}{m^*_k+1} + 2\delta.
    \end{equation}
    As a result, since $t \not\in \calB_j$, we have
    \begin{equation}
        \frac{\tilde{\mu}_{j,k}(t)}{z_0} < \frac{\mu_k + \delta}{z^* - \delta} < \frac{\mu_k + \delta}{\frac{\mu_k}{m^*_k+1} + \delta} = (m^*_k+1) \cdot \frac{\mu_k + \delta}{\mu_k + (m^*_k + 1)\delta} < (m^*_k+1).
    \end{equation}
    On the other hand, according to \theoremref{thrm:PNE}, we have $z^* \le \mu_k / m^*_k$. As a result,
    \begin{equation}
        \frac{\tilde{\mu}_{j,k}(t)}{z_0} > \frac{\mu_k - \delta}{z^* - \delta} > \frac{\mu_k - \delta}{\frac{\mu_k}{m^*_k} - \delta} = m^*_k \cdot \frac{\mu_k - \delta}{\mu_k - m_k^*\delta} > m_k^*.
    \end{equation}
    Combining the above two equations and we get for any $k \in \tilde{\calM}_j(t)$,
    \begin{equation}
        \floor{\frac{\tilde{\mu}_{j,k}(t)}{z_0}} = m_k^*.
    \end{equation}
    As a result, $\tilde{h}_{j,t}(z_0) \ge \sum_{k \in \calM^*} m_k^* = N$. Therefore, $\tilde{z}_j(t) \ge z_0$, and
    \begin{equation}
        \tilde{m}_{j,k}(t) = \floor{\frac{\tilde{\mu}_{j,k}(t)}{ \tilde{z}_{j,k}(t)}} \le \floor{\frac{\tilde{\mu}_{j,k}(t)}{z_0^*}} = m_k^*, \quad \forall k \in \tilde{\calM}_j(t).
    \end{equation}
    Since $m_k^* = 0$ for all $ k \in \tilde{\calM}_j(t) \backslash \calM$, we have $\tilde{\calM}_j(t) = \calM^*$ and $\forall k \in \calM^*$, $\floor{\tilde{\mu}_{j,k}(t) / z_0} = \floor{\tilde{\mu}_{j,k}(t) / \tilde{z}_j(t)}$. Now we can conclude that $\tilde{\bfm}^*_j(t) = \bfm^*$.
    
    Furthermore, let $k, k' \in \calM^*$ ($k \ne k'$) and suppose $\mu_k / m^*_k < r_k < r_{k'} = \mu_{k'} / m^*_{k'}$. Then by the choice of $\delta$, we have $\mu_k' / m_{k'}^* - \mu_k / m_k^* > 2\delta$. As a result, since $t \not\in \calB_j$, we have
    \begin{equation}
        \tilde{r}_{j,k}(t) - \tilde{r}_{j,k'}(t) = \frac{\tilde{\mu}_{j,k}(t)}{\tilde{m}_{j,k}(t)} - \frac{\tilde{\mu}_{j,k'}(t)}{\tilde{m}_{j,k'}(t)} = \frac{\tilde{\mu}_{j,k}(t)}{m^*_k} - \frac{\tilde{\mu}_{j,k'}(t)}{m^*_{k'}} \ge \frac{\mu_k - \delta}{m^*_k} - \frac{\mu_{k'} + \delta}{m^*_{k'}} > \left(2 - \frac{1}{m^*_k} - \frac{1}{m^*_{k'}}\right)\delta > 0.
    \end{equation}
    This equation indicates that we could get the same order if sorting $\{r^*_k = \mu_k / m^*_k\}_{k \in \calM^*}$ and $\{\tilde{r}_{j,k}(t)\}_{k \in \calM^*}$ by descending order. Together with the conclusion $\tilde{\bfm}_j(t) = \bfm^*$, we can get $\tilde{l}_j(t) = l^*$.
    
    (2) We now show that $|\tilde{r}_{j,k_0}(t) - z^*| \le \delta$ where $k_0$ is the last element in $\tilde{l}_j(t)$.
    
    From the proof of part (1), we know that $z^* - \delta \le \tilde{z}_j(t)$. In addition, since $t \not\in \calB_j$, we have
    \begin{equation}
        \tilde{z}_j(t) = \tilde{r}_{j,k_0}(t) = \frac{\tilde{\mu}_{j,k_0}(t)}{\tilde{m}_{j,k_0}(t)} < \frac{\mu_{k_0} + \delta}{m^*_{k_0}} < \frac{\mu_{k_0}}{m^*_{k_0}} + \delta = z^* + \delta.
    \end{equation}
    Here the last equation is based on the fact that $l^* = \tilde{l}(t)$ and hence $z^* = \min_{k' \in \calM^* }\mu_{k'} / m^*_{k'} = \mu_{k_0} / m^*_{k_0}$. Now the claim follows.
\end{proof}

\subsubsection{Proof of \lemmaref{lemma:subseteq}} \label{sect:proof-lemma-subseteq}
\begin{proof}
    We only need to show that $\calA_j \subseteq \calB_j \cup \calC_j \cup \calD_j$. Let $t \in \calA_j \backslash (\calB_j \cup \calC_j)$.
    
    Because $t \in \calA_j$, we have that there exists $k \in \calM^* \backslash \tilde{\calM}_j(t)$. Since
    \begin{equation}
        \left\{\calM^* \backslash \tilde{\calM}_j(t) \ne \emptyset\right\} = \left\{\tilde{\calM}_j(t) \backslash \calM^* \ne \emptyset\right\} \cup \left\{\tilde{\calM}_j(t) \subsetneq \calM^* \right\}.
    \end{equation}
    We consider these two cases respectively.
    
    (1) Consider the $\calM^* \backslash \tilde{\calM}_j(t) \ne \emptyset$ case.
    
    Let $k' \in \tilde{\calM}_j(t) \backslash \calM^*$. Therefore, we have $\tilde{\mu}_{j,k}(t) \le \tilde{\mu}_{j,k'}(t)$. Otherwise, a player deviate from arm $k'$ to arm $k$ will increase the reward, which indicates that $\tilde{\bfm}_j(t)$ is not a Nash equilibrium \textit{w.r.t.} $\{\tilde{\mu}_{j,k}(t)\}_{k=1}^K$ and leads to a contradiction.
    
    In addition, since $t \not\in \calB_j$, we have that $|\tilde{\mu}_{j,k'}(t) - \mu_{k'}| < \delta$. Furthermore, by the definition of $\delta_0$ and the property of the Nash equilibrium $\bfm^*$ \textit{w.r.t.} $\{\mu_{k}\}_{k=1}^K$, we have
    \begin{equation}
        \frac{\mu_k}{m^*_k} - \mu_{k'} \ge \delta_0 > 2 \delta.
    \end{equation}
    As a result,
    \begin{equation}
        \mu_k - \tilde{\mu}_{j,k}(t) \ge m_k^*(\mu_{k'} + 2\delta) - \tilde{\mu}_{j, k'}(t) \ge 2\delta + (\mu_{k'} - \tilde{\mu}_{j,k'}(t)) > \delta.
    \end{equation}
    Therefore, $t \in \calD_j$.
    
    (2) Consider the $\tilde{\calM}_j(t) \subsetneq \calM^*$ case.
    
    Since $\tilde{\calM}_j(t) \subsetneq \calM^*$, there exists $k' \in \tilde{\calM}_j(t)$ such that $\tilde{m}_{j,k'}(t) > m_{k'}^*$. Because $k \not\in \tilde{\calM}_j(t)$, we have $\tilde{\mu}_{j,k}(t) \le \tilde{\mu}_{j, k'}(t) / \tilde{m}_{j,k'}(t)$. Furthermore, by the definition of $\delta_0$ and the property of the Nash equilibrium $\bfm^*$ \textit{w.r.t.} $\{\mu_{k}\}_{k=1}^K$, we have
    \begin{equation}
        \mu_k \ge \frac{\mu_{k'}}{m^*_{\mu_{k'}} + 1} + 2\delta \ge \frac{\mu_{k'}}{\tilde{m}_{j,k'}(t)} + 2\delta.
    \end{equation}
     In addition, since $t \not\in \calB_j$, we have that $|\tilde{\mu}_{j,k'}(t) - \mu_{k'}| < \delta$. As a result,
     \begin{equation}
         \mu_k - \tilde{\mu}_{j,k}(t) \ge \frac{\mu_{k'}}{\tilde{m}_{j,k'}(t)} + 2\delta - \tilde{\mu}_{j,k}(t) > \frac{\tilde{\mu}_{j,k'}(t) - \delta}{\tilde{m}_{j,k'}(t)} + 2\delta - \tilde{\mu}_{j,k}(t) \ge \delta\left(2 - \frac{1}{\tilde{m}_{j,k'}(t)}\right) \ge \delta.
     \end{equation}
     Therefore, $t \in \calD_j$.
     
     Now the claim follows.
\end{proof}

\subsubsection{Proof of \lemmaref{lemma:subset-expectation}} \label{sect:proof-lemma-subset-expectation}
\begin{proof}
    Since $\bbE[|\calB_j \cup \calC_j \cup \calD_j|] \le \bbE[|\calB_j|] + \bbE[|\calC_j|] + \bbE[|\calD_j|]$, we provide the upper bounds of $\bbE[|\calB_j|]$, $\bbE[|\calC_j|]$, and $\bbE[|\calD_j|]$, respectively. Let $\calF_t$ be the $\sigma$-algebra generated by $\left\{X_{k}(s), \forall k \in [K]\right\}_{s=1}^t$.
    
    (1) We first show that $\bbE[|\calB_j|] \le NK(17 + 4\delta^{-2})$. 
    
    For any $k \in [K]$, let $\calB_{j,k} = \left\{t > K': k \in \tilde{\calM}_j(t), |\tilde{\mu}_{j,k}(t) - \mu_k| \ge \delta\right\}$. Let $1 \le p_k(t) \le N$ be the index of the last element that is $k$ in $\tilde{l}_t$. Define
    \begin{equation}
        \calB_{j,k,1} = \left\{t \in \calB_{j,k}: t \equiv p_k(t) (\mo N) \right\}.
    \end{equation}
    
    Let $H = \left\{t > K': t \equiv p_k(t) (\mo N)\right\}$ and $C_t = \bbI[\pi_j(t) = k]$. Note that $\bbI[t \in H]$ is $\calF_{t-1}$-measurable and $\bbP[C_t = 1 | t \in H] \ge 1 / 2$ by the design of the algorithm. According to \lemmaref{lemma:number-rounds}, we can get that
    \begin{equation}
        \bbE[|\calB_{j,k,1}|] \le \sum_{t > K'} \bbP\left[t \in H, |\hat{\mu}_{j,k}(t) - \mu_k| \ge \delta\right] \le 4(4 + \delta^{-2}).
    \end{equation}
    
    $\forall t \in H$, it is always the last possible time to choose arm $k$ in the corresponding block \text{w.r.t.} $t$. As a result, $\forall t \in H$, $\hat{\mu}_{j,k}(t)$ is the estimations adopted in the next block. Formally, $\forall s > N$ and $s \in \calB_{j,k}$, we have $\tilde{\mu}_{j,k}(s) = \hat{\mu}_{j, k}(\max\{t \in H: t < s\})$ and $\max\{t \in H: t < s\} \in \calB_{j,k,1}$. In addition, each $t \in \calB_{j,k,1}$ corresponds to at most $N$ elements in $\calB_{j,k}$, which means that
    \begin{equation}
        \bbE[|\calB_{j,k}|] \le N + N \cdot \bbE[|\calB_{j,k,1}|] \le N(17 + 4\delta^{-2}).
    \end{equation}
    Now by the union bound, we have
    \begin{equation} \label{eq:proof-expectation-b}
        \bbE[|\calB_j|] \le \sum_{k=1}^K \bbE[|\calB_{j,k}|] \le NK(17 + 4\delta^{-2}).
    \end{equation}
    
    (2) We then show that $\bbE[|\calC_j|] \le 60N^2$.
    
    For any $k \in \calM^*$, let $\calC_{j,k} = \left\{t > K': \tilde{b}_{j,k}(t) < \mu_k\right\}$. By the design of our algorithm, $\tilde{b}_{j,k}(t)$ is updated at the start of each block. As a result
    \begin{equation}
        \bbE[|\calC_{j,k}|] = \sum_{t  > K'}\bbP\left[\tilde{b}_{j,k}(t) < \mu_k\right] \le N\sum_{t \ge 0}\bbP\left[\tilde{b}_{j,k}\left(tN + 1\right) < \mu_k\right] \le N + N\sum_{t \ge 1} \bbP\left[\hat{b}_{j,k}(tN) < \mu_k\right].
    \end{equation}
    According to \lemmaref{lemma:garivier} (Theorem 10 in \citep{garivier2011kl}), we can get that for all $t \ge 3$,
    \begin{equation}
        \bbP\left[\hat{b}_{j,k}\left(tN\right) < \mu_k\right] \le \ceil{f(tN)\log f(tN)}e^{1-f(tN)}.
    \end{equation}
    Here $f(tN) = \log (tN) + 4\log (\log (tN))$. 
    As a result,
    \begin{equation}
        \begin{aligned}
            \bbE[|\calC_{j,k}|] & \le 3N + N \sum_{t \ge 3} \ceil{f(tN)\log f(tN)}e^{1-f(tN)} \\
            & \le 3N + N \sum_{t \ge 3}\ceil{f(s)\log f(s)}e^{1-f(s)} \le 60N,
        \end{aligned}
    \end{equation}
    where the last inequality is based on the proof of Lemma 4 in \citep{wang2020optimal}. Therefore, by the union bound, we have
    \begin{equation} \label{eq:proof-expectation-c}
        \bbE[|\calC_j|] \le \sum_{k \in \calM^*}\bbE[|C_{j,k}|] \le 60N^2.
    \end{equation}
    
    (3) Finally, we show that $\bbE[|\calD_j|] \le N^2K(17K + 4\delta^{-2})$.
    
    For any $k \in \calM^*$, let
    \begin{equation}
        \calD_{j,k} = \left\{t \in \calA_j \backslash (\calB_j \cup \calC_j): k \not\in \tilde{\calM}_j(t) \text{ and } |\tilde{\mu}_{j,k}(t) - \mu_k| \ge \delta\right\}.
    \end{equation}
    Fix $k \in \calM^*$. Suppose $t \in \calD_{j,k}$. Since $t \not\in C_j$, we have $\tilde{b}_{j,k}(t) \ge \mu_k$. In addition, since $t \not\in \calB_j$, we have that for any $k' \in \tilde{\calM}_j(t)$, $|\tilde{\mu}_{j,k'}(t) - \mu_{k'}| < \delta$. Let $k_0$ be the last element in this list $\tilde{l}_j(t)$.
    
    In addition, because $t \in \calA_j$, we have $\calM^* \backslash \tilde{\calM}_j(t) \ne \emptyset$. Because $\left\{\calM^* \backslash \tilde{\calM}_j(t) \ne \emptyset\right\} = \left\{\tilde{\calM}_j(t) \backslash \calM^* \ne \emptyset\right\} \cup \left\{\tilde{\calM}_j(t) \subsetneq \calM^* \right\}$. We show that $k \in \calH_{j}(t)$ in these two cases, respectively.
    
    (a) Suppose $\tilde{\calM}_j(t) \backslash \calM^* \ne \emptyset$. Let $k'$ be any element in $\tilde{\calM}_j(t) \backslash \calM^*$. As a result, since $k \in \calM^*$ and $k' \not\in \calM^*$, under \assumptionref{assum:unique}, we have $\mu_k \ge \mu_{k'} + 2\delta$. In addition, by the construction of $\tilde{l}_j(t)$, we have $\tilde{\mu}_{j, k'}(t) \ge \tilde{r}_{j, k'}(t) \ge \tilde{r}_{j,k_0}(t)$. As a result,
    \begin{equation}
        \tilde{b}_{j,k}(t) \ge \mu_k \ge \mu_{k'} + 2 \delta \ge \tilde{\mu}_{j,k'}(t) + \delta > \tilde{r}_{j,k_0}(t).
    \end{equation}
    As a result, $k \in \calH_j(t)$ by the construction of the algorithm.

    (b) Suppose $\tilde{\calM}_j(t) \subsetneq \calM^*$. There exists $k' \in \tilde{\calM}_j(t)$ such that $\tilde{m}_{j,k'}(t) > m_{k'}^*$. Because $k \not\in \tilde{\calM}_j(t)$, we have $\tilde{\mu}_{j,k}(t) \le \tilde{\mu}_{j, k'}(t) / \tilde{m}_{j,k'}(t)$. Furthermore, by the definition of $\delta_0$ and the property of the Nash equilibrium $\bfm^*$ \textit{w.r.t.} $\{\mu_{k}\}_{k=1}^K$, we have
    \begin{equation}
        \mu_k \ge \frac{\mu_{k'}}{m^*_{\mu_{k'}} + 1} + 2\delta \ge \frac{\mu_{k'}}{\tilde{m}_{j,k'}(t)} + 2\delta.
    \end{equation}
    As a result, by the construction of $\tilde{l}_j(t)$, we have $\tilde{\mu}_{j,k'}(t) / \tilde{m}_{j,k'}(t) \ge \tilde{r}_{j,k_0}(t)$. Therefore, we can get
    \begin{equation}
        \tilde{b}_{j,k}(t) \ge \mu_k \ge \frac{\mu_{k'}}{\tilde{m}_{j,k'}(t)} + 2\delta \ge \frac{\tilde{\mu}_{j,k'}(t) - \delta}{\tilde{m}_{j,k'}(t)} + 2\delta \ge \tilde{r}_{j,k_0}(t) + \delta\left(2 - \frac{1}{\tilde{m}_{j,k'}(t)}\right) > \tilde{r}_{j,k_0}(t).
    \end{equation}

    To conclude, from part (a) and (b), we get that $k \in \calH_j(t)$. Therefore, when $t_0 = N$ in \algorithmref{alg:SMAA}, arm $k$ will be chosen with probability at least $1 / (2K)$. Now use the similar techniques in the proof of part (1). Let
    \begin{equation}
        \calD_{j, k, 1} = \left\{t \in \calD_{j,k}: (t + j) \equiv N - 1 (\mo N)\right\}.
    \end{equation}
    By the construction of the algorithm, for each $t \in \calD_{j,k,1}$, arm $k$ will be chosen with probability at least $1 / (2K)$. Let $H = \{t \in \calA_j \backslash (\calB_j \cup \calC_j): k \not\in \tilde{\calM}_j(t) \text{ and } (t + j) \equiv N - 1 (\mo N)\}$, $C_t = \bbI[\pi_j(t) = k]$. According to \lemmaref{lemma:number-rounds}, we have
    \begin{equation}
        \bbE[|\calD_{j,k,1}|] = \sum_{t \ge 1} \bbP[t \in \calH, |\tilde{\mu}_{j,k}(t) - \mu_k| \ge \delta] \le 16K^2 + 4K\delta^{-2}.
    \end{equation}
    Similar to the proof of part (1), we have that
    \begin{equation}
        \bbE[|\calD_{j,k}|] \le N(1 + \bbE[|\calD_{j,k,1}|]) \le NK(17K + 4\delta^{-2}).
    \end{equation}
    Finally, by the union bound, we have
    \begin{equation} \label{eq:proof-expectation-d}
        \bbE[|\calD_j|] \le \sum_{k \in \calM^*}\bbE[|\calD_{j,k}|] \le N^2K(17K + 4\delta^{-2}).
    \end{equation}
    
    Now the claim follows by combining \equationsref{eq:proof-expectation-b}, \eqref{eq:proof-expectation-c}, and \eqref{eq:proof-expectation-d}.
\end{proof}

\subsubsection{Proof of \lemmaref{lemma:bad-arms}} \label{sect:proof-lemma-bad-arms}
\begin{proof}
    Define $w(t) = \sum_{s=1}^t \bbI[s \in \calG_{j,k}]$ and $s_0 = (\log T + 4\log (\log T)) / \kl(\mu_k + \delta, z^* - \delta)$ where $z^*$ is provided in \equationref{eq:z-star}. Further, let
    \begin{equation}
        \calG_{j,k,1} = \left\{t \in \calG_{j,k}: \left|\tilde{\mu}_{j,k}(t) - \mu_k\right| \ge \delta\right\} \quad \text{ and } \quad \calG_{j,k,2} = \{t \in \calG_{j,k}: w(t) < s_0 + 1\}.
    \end{equation}
    
    (1) We first show that $\calG_{j,k} \subseteq (\calG_{j,k,1} \cup \calG_{j,k,2})$.
    
    Suppose $\calG_{j,k} \backslash (\calG_{j,k,1} \cup \calG_{j,k,2}) \ne \emptyset$. Let $t \in \calG_{j,k} \backslash (\calG_{j,k,1} \cup \calG_{j,k,2})$. Since $t \in \calG_{j,k}$, we have that $t \not\in \calA_j$ and $\calM^* = \tilde{\calM}_j(t)$. As a result, $k \not\in \tilde{\calM}_j(t)$. Therefore, by the design of \algorithmref{alg:SMAA}, arm $k$ will be chosen at most one time in the block corresponding to round $t$. As a result, $\tilde{\tau}_{j,k}(t) \ge w(t) - 1$. In addition, since $t \not\in \calG_{j,k,2}$, we have $w(t) \ge s_0 + 1$. Hence, we have
    \begin{equation} \label{eq:proof-lemma-bad-arms-1}
        \tilde{\tau}_{j,k}(t) \ge s_0.
    \end{equation}
    
    Let $k' = \tilde{l}_{j, N}(t) \in \calM^*$, \textit{i.e.}, the last element in this list $\tilde{l}_j(t)$ with the smallest estimated average reward $\tilde{r}_{j,k'}(t)$. We note that $\pi_j(t) = k$ only when $\tilde{b}_{j,k}(t) \ge \tilde{r}_{j,k'}(t)$ by the choice of $\calH_j(t)$ in Line 9 of \algorithmref{alg:SMAA}. In addition, since $t \not\in (\calA_j \cup \calB_j)$, according to \lemmaref{lemma:clean-event}, we have that
    \begin{equation} \label{eq:proof-lemma-bad-arms-2}
        \tilde{b}_{j,k}(t) \ge \tilde{r}_{j,k'}(t) \ge z^* - \delta.
    \end{equation}
    Furthermore, since $t \not\in \calG_{j,k,1}$, we have $|\tilde{\mu}_{j,k}(t) - \mu_k| < \delta$. Because $k \not\in \calM^*$, under \assumptionref{assum:unique}, we have $\mu_k < z^*$ (Otherwise, a player that chooses arm $q = k^*_{|\calM^*|}$ deviate to choose arm $k$ will increase the reward). As a result, by the construction of $\delta$, we have $z^* - \mu_k > 2 \delta$. Hence,
    \begin{equation} \label{eq:proof-lemma-bad-arms-3}
        \tilde{\mu}_{j,k}(t) \le \mu_k + \delta < z^* - \delta.
    \end{equation}
    As a result, we can get that
    \begin{equation}
        s_0 \kl\left(\tilde{\mu}_{j,k}(t), z^* - \delta\right) \le \tilde{\tau}_{j,k}(t)\kl\left(\tilde{\mu}_{j,k}(t), z^* - \delta\right) \le \tilde{\tau}_{j,k}(t)\kl\left(\tilde{\mu}_{j,k}(t), \tilde{b}_{j,k}(t)\right) \le \log T + 4 \log (\log T).
    \end{equation}
    Here the first inequality is from \equationref{eq:proof-lemma-bad-arms-1}. The second inequality stems from \equationref{eq:proof-lemma-bad-arms-2} and the fact that $\forall 0 < x < 1$, $y \mapsto \kl(x, y)$ is an increasing function when $x < y < 1$ (guaranteed by \equationref{eq:proof-lemma-bad-arms-3}). The last inequality is according to the definition of $\tilde{b}_{j,k}(t)$. By the definition of $s_0$, we can get that
    \begin{equation}
        \kl\left(\tilde{\mu}_{j,k}(t), z^* - \delta\right) \le \kl\left(\mu_k + \delta, z^* - \delta\right).
    \end{equation}
    As a result, $\tilde{\mu}_{j,k}(t) \ge \mu_k + \delta$ by the fact that $x < y < 1$, $x \mapsto \kl(x, y)$ is a decreasing function when $0 < x < 1$. As a result, $t \in \calG_{j,k,1}$, which leads to a contradiction.
    
    (2) We show that $\bbE[|\calG_{j,k,1}|] \le 5 + 2\delta^{-2}$.
    
    Let $H = \{t \in \calG_{j,k}: |\hat{\mu}_{j,k}(t) - \mu_k| \ge \delta\}$. According to \lemmaref{lemma:number-rounds} with $C_t = c = 1$, we have that $\bbE[H] \le 4 + 2\delta^{-2}$. Since $k \not\in \calM^* = \tilde{\calM}_j(t)$, arm $k$ can be chosen at most one time in the corresponding block \textit{w.r.t.} $t$. As a result, $\bbE[|\calG_{j,k,1}|] \le \bbE[|H|] \le 5 + 2\delta^{-2}$.
    
    (3) We show that $\bbE[|\calG_{j,k,2}|] \le (\log T + 4\log (\log T)) / \kl(\mu_k + \delta, z^* - \delta)$.
    
    By the construction of $w(t)$, we have $\bbE[|\calG_{j,k,2}|] \le s_0$.
    
    Finally, we can prove that
    \begin{equation}
        \bbE[|\calG_{j,k}|] \le \bbE[|\calG_{j,k,1}|] + \bbE[\calG_{j,k,2}] \le \frac{\log T + 4 \log (\log T)}{\kl(\mu_k + \delta, z^* - \delta)} + 5 + 2\delta^{-2}.
    \end{equation}
\end{proof}

\subsection{Proofs in \sectionref{sect:additional}} \label{sect:proof-additional}
\subsubsection{Proof of \propositionref{prop:justification-assumption}} \label{sect:proof-assumption}
\begin{proof}
    Let $A$ denote the event that \assumptionref{assum:unique} holds and $B$ denote the event that for all $i, j \in [K]$ and $i \ne j$, $\mu_i / \mu_j$ is an irrational number. Let $B_{ij}$ denote the event that $\mu_i / \mu_j$ is an irrational number. It is obvious that
    \begin{equation}
        \bigcap_{1 \le i <  j \le K}B_{ij} = B \subseteq A.
    \end{equation}
    As a result,
    \begin{equation} \label{eq:proof-justification-rational-1}
        \bbP(A) \ge \bbP(B) = 1 - \bbP(\bar{B}) = 1 - \bbP\left(\bigcup_{1 \le i < j \le K}\bar{B}_{i,j}\right) \ge 1 - \sum_{1 \le i < j \le K}\bbP(\bar{B}_{ij}).
    \end{equation}
    In addition,
    \begin{equation} \label{eq:proof-justification-rational-2}
        \bbP(\bar{B}_{ij}) = \bbE[\bbP(\mu_j / \mu \text{ is rational}) | \mu_i = \mu].
    \end{equation}
    For a fixed $\mu \in (0, 1]$, since rational number is countable, let $a_1, a_2, a_3, \dots, $ be the rational numbers in $(0, 1 / \mu]$. Fix an $\epsilon > 0$ and define the following intervals $[l_1, u_1], [l_2, u_2], \dots$ with
    \begin{equation}
        l_t = a_t - \epsilon / 2^t \quad \text{and} \quad r_t = a_t + \epsilon / 2^t.
    \end{equation}
    As a result,
    \begin{equation}
        \bbP(\mu_j / \mu \text{ is rational}) = \bbP(\mu_j / \mu \in \{a_1, a_2, \dots\}) \le \bbP\left(\mu_j / \mu \in \bigcup_{t \ge 1} [l_t, u_t]\right) \le \sum_{t \ge 1}\bbP\left(\mu_j / \mu \in [l_t, u_t]\right).
    \end{equation}
    Since the density ratio of distribution $D$ is bounded by $M$, we have
    \begin{equation}
        \sum_{t \ge 1}\bbP\left(\mu_j / \mu \in [l_t, u_t]\right) \le \sum_{t \ge 1} \bbP\left(\mu_j \in [l_t \mu, u_t \mu]\right) \le M\mu \sum_{t \ge 1} \epsilon / 2^{t-1} \le 2M\epsilon.
    \end{equation}
    By letting $\epsilon \rightarrow 0$, we have $\bbP(\mu_j / \mu \text{ is rational}) = 0$. Now combining \equationsref{eq:proof-justification-rational-1} and \eqref{eq:proof-justification-rational-2}, we have $\bbP(A) \ge 1$. Now the claim follows.
\end{proof}

\subsubsection{Proof of \propositionref{prop:strong-pne}} \label{sect:proof-strong-pne}
\begin{proof}
    Let $\bfs$ be a Nash equilibrium in \theoremref{thrm:PNE}. Suppose there exists a beneficial deviation for players $B \in [N]$ and the corresponding strategies $\bfs'_B$.

    We first prove that $\bfs_B' \subseteq \calM^*$. Suppose there exists a player $j \in B$ and $s'_j \not\in \calM^*$. Then $\calU_j^{\single}(\bfs'_B, \bfs_{-B}) = \mu_{s'_j} < z^*$ by \theoremref{thrm:PNE}. However, since the original profile $\bfs$ is a Nash equilibrium, we have $\calU_j^{\text{single}}(\bfs) \ge z^*$. This leads to the contradiction with the definition of beneficial deviation. As a result, $\bfs_B' \subseteq \calM^*$.

    We now show that $\bfm(\bfs_B', \bfs_{-B}) = \bfm(\bfs)$. Suppose $\bfm(\bfs_B', \bfs_{-B}) \ne \bfm(\bfs)$, there must exist $k \in \calM^*$ such that $m_k(\bfs'_B, \bfs_{-B}) > m_k(\bfs) = m_k^*$. As a result, there must exist a player $j \in B$ such that $s'_j = k$ and $\calU_j^{\single}(\bfs'_B, \bfs_{-B}) = \mu_k / m_k(\bfs'_B, \bfs_{-B}) < \mu_k / m_k^* < z^*$. However, since the original profile $\bfs$ is a Nash equilibrium, we have $\calU_j^{\text{single}}(\bfs) \ge z^*$. This leads to the contradiction with the definition of beneficial deviation. As a result, $\bfm(\bfs_B', \bfs_{-B}) = \bfm(\bfs)$.

    Finally, we show that such beneficial deviation does not exist. Base on the previous two claims, we have $\bfm(\bfs_B', \bfs_{-B}) = \bfm(\bfs)$. As a result, the rewards earned by the players in $B$ after deviation are a permutation of the original rewards earned by them. Therefore,
    \begin{equation}
        \sum_{j \in B} \calU^{\single}_j(\bfs_B', \bfs_{-B}) = \sum_{j \in B} \calU^{\single}_j(\bfs),
    \end{equation}
    which contradicts to the definition of the beneficial deviation which requires that at least one inequality holds strictly for players in $B$. Now the claim follows.
\end{proof}

\subsubsection{Proof of \propositionref{prop:mne-vs-pne}} \label{sect:proof-mne-vs-pne}
\begin{proof}
    Since $\bfsigma$ is a symmetric mixed Nash equilibrium, we suppose $\bfsigma_1 = \bfsigma_2 = \dots = \bfsigma_N = \bfp = (p_1, p_2, \dots, p_K)$ where $p_k$ represents the probability for each player to pull arm $k$. Let $\calP = \{k: p_k > 0\}$ be the set of all arms that can be chosen according to $\bfp$. It is obvious that, if $\calP \subseteq \calM^*$, we must have $W(\bfsigma) \le W^{\text{PNE}}$. Now we consider the $\calP \backslash \calM^* \neq \emptyset$ case.
    
    Since $\bfsigma$ is a Mixed Nash equilibrium, we have
    \begin{equation} 
         \bbE_{\bfs \in \bfsigma}\left[\calU_j^{\single}(\bfs)\right] = \sum_{k \in \calP}p_k\bbE_{\bfs \in \bfsigma}\left[\calU_j^{\single}(k, \bfs_{-j})\right] \leq \sum_{k \in \calP}p_k\bbE_{\bfs \in \bfsigma}\left[\calU_j^{\single}(\bfs)\right] = \bbE_{\bfs \in \bfsigma}\left[\calU_j^{\single}(\bfs)\right].
    \end{equation}
    This indicates that all the inequalities should be equations in the second step and we have that there exists a constant $C > 0$, such that
    \begin{equation} \label{eq:proof-mne-0}
        \forall k \in \calP, \quad \bbE_{\bfs \in \bfsigma}\left[\calU_j^{\single}(k, \bfs_{-j})\right] = C.
    \end{equation}
    Now we analyze the $\bbE_{\bfs \in \bfsigma}\left[\calU_j^{\single}(k, \bfs_{-j})\right]$ term for any $k \in \calP$. Let $\bfm_{-j}(\bfs) = \{m_{-j, k}(\bfs)\}_{k=1}^K$, where $m_{-j, k}(\bfs)$ denotes the number of players except player $j$ that choose arm $k$. Then $\bbE_{\bfs \in \bfsigma}\left[\calU_j^{\single}(k, \bfs_{-j})\right]$ can be expanded as follows:
    \begin{equation} \label{eq:proof-mne-2}
        \bbE_{\bfs \in \bfsigma}\left[\calU_j^{\single}(k, \bfs_{-j})\right] = \sum_{t=0}^{N-1}\frac{\mu_k}{t + 1}\bbP[m_{-j, k}(\bfs) = t] = \mu_k \sum_{t=0}^{N-1}\frac{1}{t+1}\binom{N-1}{t}p_k^t(1-p_k)^{N - 1 - t}.
    \end{equation}
    To calculate the above equation, we define the following generating function,
    \begin{equation}
        F(x) = (p_k x + (1-p_k))^{N-1} = \sum_{t=0}^{N-1}a_t x^t,
    \end{equation}
    where $a_t = \binom{N-1}{t}p_k^t(1-p_k)^{N - 1 - t}$. As a result, \equationref{eq:proof-mne-2} can be calculated as follows.
    \begin{equation} \label{eq:proof-mne-3}
        \bbE_{\bfs \in \bfsigma}\left[\calU_j^{\single}(k, \bfs_{-j})\right] = \mu_k \int_0^1F(x)\mathrm{d}x = \mu_k \cdot \frac{(p_k x + (1-p_k))^N}{p_kN}\Bigg|_{x=0}^1 = \frac{\mu_k\left(1 - (1-p_k)^N\right)}{p_kN} = C.
    \end{equation}
    Hence, $\mu_k\left(1 - (1-p_k)^N\right) = C p_k N$ and
    \begin{equation}
        W(\bfsigma) = \sum_{k=1}^K \mu_k \bbP_{\bfs \sim \bfsigma}[m_k(\bfs) > 0] = \sum_{k=1}^K \mu_k \left(1 - (1 - \mu_k)^N\right) = \sum_{k \in \calP} C p_k N = CN.
    \end{equation}
    In addition, \equationref{eq:proof-mne-3} indicates that
    \begin{equation} \label{eq:proof-mne-4}
        \frac{1 - (1-p_k)^N}{p_k} = \frac{CN}{\mu_k}.
    \end{equation}
    It is easy to check that function $(1 - (1 - p_k)^N) / p_k$ is non-increasing and is bounded in $[1, N]$ when $0 \le p_k \le 1$. As a result,
    \begin{equation}
        \forall k \in \calP, \quad \frac{CN}{\mu_k} \le N.
    \end{equation}
    Hence we have $CN \le \mu_k N$ for any $k \in \calP$. Combining \equationref{eq:proof-mne-4}, we have
    \begin{equation}
        W(\bfsigma) \le N \mu_k, \quad \text{for any } k \in \calP.
    \end{equation}
    Since we consider $\calP \backslash \calM^*$ case, there exists an arm $k \in \calP \backslash \calM^*$. Since it is not in the set $\calM^*$, according to \theoremref{thrm:PNE}, we have $\mu_k < z^*$. As a result,
    \begin{equation}
        W(\bfsigma) \le N \mu_k \le N z^* = \sum_{k \in \calM^*}m_k^*z^* \le \sum_{k \in \calM^*}\mu_k =   W^{\text{PNE}}.
    \end{equation}
    Now the claim follows.
\end{proof}

\subsubsection{Proof of \propositionref{prop:regret-new}} \label{sect:proof-regret-new}
\begin{proof}
    Consider the set $\calJ$ defined in \equationref{eq:time-sets}, according to \lemmaref{lemma:subseteq} and \lemmaref{lemma:subset-expectation}, we have
    \begin{equation}
        \bbE[|\calJ|] \le \sum_{j=1}^N \bbE[|\calA_j \cup \calB_j|] \le \sum_{j=1}^N \bbE[|\calB_j \cup \calC_j \cup \calD_j|] \le 8N^3K(12K + \delta^{-2}).
    \end{equation}
    Since $\bbE[|\calJ|] = O(1)$, the expected number of blocks that contain rounds $t \in \calJ$ is also in $O(1)$. As a result, the differences between $\bbE[\regret_j(T)]$ (defined in \equationref{eq:regret}) and $\bbE[\regret_j'(T)]$ in these blocks are also $O(1)$.
    
    Now we analyze the differences between the two regrets for the blocks of which all rounds are not in $\calJ$. Suppose a block with rounds $vN, vN + 1, vN+2, \dots v\cdot(N+1)$ for an integer $v$ and $(vN + s) \not\in \calJ$ for any $1 \le s \le N$. According to \lemmaref{lemma:clean-event}, each player $j$ at these rounds calculates the correct $\tilde{l}_j(t) = l^*$. In addition, at most one player deviates according to \algorithmref{alg:SMAA} to explore sub-optimal arms. As a result, the optimal arm for player $j$ at each round in the block is exactly the arm that he plans to pull without the exploration in Lines 16-17 in \algorithmref{alg:SMAA}. Formally, for each $t \in \{vN, vN + 1, \dots, v\cdot(N+1)\}$,
    \begin{equation}
        \max_{k \in [K]}\frac{\mu_k}{M_k(t) + \bbI[\pi_j(t) \ne k]} = \frac{\mu_k}{m_k^*}.
    \end{equation}
    Since agent $j$ calculates the correct Nash equilibrium and $\tilde{l}_j(t) = l^*$, each arm $k \in \calM^*$ is chosen for $m_k^*$ times in $\tilde{l}_j(t)$. As a result,
    \begin{equation}
        \sum_{t=vN+1}^{v\cdot (N+1)}\max_{k \in [K]}\frac{\mu_k}{M_k(t) + \bbI[\pi_j(t) \ne k]} = \sum_{k \in \calM^*} \frac{\mu_k}{m_k^*} \cdot m_k^* = \sum_{k \in \calM^*} \mu_k = N \cdot r^* = \sum_{t=vN+1}^{v\cdot (N+1)} r^*.
    \end{equation}
    As a result, the differences between the two regrets $\bbE[\regret_j(T)]$ and $\bbE[\regret_j'(T)]$ are $0$ in these blocks.

    Combining these results, we know that $\bbE[\regret_j(T)]$ and $\bbE[\regret_j'(T)]$ have differences at most $O(1)$. According to \theoremref{thrm:main-regret}, we have
    \begin{equation}
        \bbE[\regret_j'(T)] \le \bbE[\regret_j(T)] + O(1) \le \sum_{k \not\in \calM^*} \frac{(z^* - \mu_k)(\log T + 4 \log (\log T))}{\kl(\mu_k + \delta, z^* - \delta)} + O(1).
    \end{equation}
    Now the claim follows.
\end{proof}

\subsubsection{Proof of \propositionref{prop:justification-consistent}} \label{sect:proof-consistent}
\begin{proof}
    By the no regret condition, we have $\sum_{j=1}^N \bbE[\regret_j(T)] \le o(T^{\alpha})$. We note that each non-equilibrium round contributes at least $\delta_0$ regret since there must exist at least one player that can gain $\delta_0$ increase in reward by deviation. As a result,
    \begin{equation}
        \begin{aligned}
            & \, \bbE[\regret_j(T)] \\
            = & \, \sum_{t=1}^T\bbE\left[\sum_{j=1}^N\left(\max_{k \in [K]}\frac{\mu_{k}}{M_{k}(t) + \bbI[\pi_j(t) \ne k]} - R_j(t)\right)\right] \\
            \ge & \, \sum_{t=1}^T\bbE\left[\left.\sum_{j=1}^N\left(\max_{k \in [K]}\frac{\mu_{k}}{M_{k}(t) + \bbI[\pi_j(t) \ne k]} - R_j(t)\right)\right| \exists k \in [K], M_k^*(t) \ne m_k^*\right] \bbP\left[\exists k \in [K], M_k^*(t) \ne m_k^*\right] \\
            \ge & \,  \delta_0 \noneq(T).
        \end{aligned}
    \end{equation}
    As a result, $\noneq(T) \le o(T^\alpha)$. Therefore, for all $k \in \calM^*$,
    \begin{equation}
        \begin{aligned}
            \bbE\left[\sum_{j=1}^N \tau_{j,k}(T)\right] \ge m_k^* \left(T - \bbE[\noneq(T)]\right).
        \end{aligned}
    \end{equation}
    Therefore, $m_k^*T - \sum_{j=1}^N \bbE[\tau_{j,k}(T)] \le o(T^\alpha)$. Now due to the fairness condition, we have that for all $\alpha > 0$, $j \in [N]$ and $k \in \calM^*$,
    \begin{equation}
        \frac{m_k^*}{N}T - \bbE[\tau_{j,k}(T)] \le o(T^\alpha).
    \end{equation}
    The claim follows.
\end{proof}

%% file: paragraphs/appendix-relaxed-setting.tex
\section{More Details about the Setting When $N$ and Rank is unknown} \label{sect:relaxed-setting}
\subsection{Algorithm Details} 
The pseudo-code of the whole method is shown in \algorithmref{alg:SMAA-relaxed} and the pseudo-code of the Musical Chairs approach is shown in \algorithmref{alg:musical-chair}. When $t \in [T_0]$, players randomly choose the arms in $[K]$ and count the number of collisions. The number of players is then estimated according to Line 10 in \algorithmref{alg:musical-chair}. When $t \in \{T_0 + 1, \dots, T\}$, each player first randomly chooses the arms in $[\hat{N}]$. If no collision occurs, the player will hold on to the arm in the remaining rounds. Otherwise, the player re-sampleso an arm uniformly in $[\hat{N}]$ and the progress goes on. The output rank of the player is the index of the arm he holds on.

\begin{algorithm}[tb]
    \caption{SMAA (without knowledge of $N$ and rank)}
    \label{alg:SMAA-relaxed}
    \begin{algorithmic}[1]
        \STATE $\hat{N}, j \leftarrow$ Musical Chairs Approach (\citep{rosenski2016multi}, \algorithmref{alg:musical-chair})
        \STATE $T_1 \leftarrow \ceil{50 K^2\log (4T)} + \ceil{\hat{N} \log (2T)}$
        \STATE Call \algorithmref{alg:SMAA} for $t$ from $T_1 + 1$ to $T$.
    \end{algorithmic}
\end{algorithm}

\begin{algorithm}[tb]
    \caption{Musical Chairs \citep{rosenski2016multi}}
    \label{alg:musical-chair}
    \begin{algorithmic}[1]
        \STATE $T_0 \leftarrow \ceil{50\log (4T) K^2}$
        \STATE $C \leftarrow 0$
        \FOR{$t \leftarrow 1$ to $T_0$}
            \STATE Sample arm $k$ uniformly in $[K]$
            \STATE Observe the collision information $\eta(t)$
            \IF{$\eta(t) = 1$}
                \STATE $C \leftarrow C + 1$
            \ENDIF
        \ENDFOR
        \STATE $\hat{N} \leftarrow \min\left\{\round\left(\frac{\log ((T_0 - C) / T_0)}{\log (1 - 1 / K)} + 1\right), K\right\}$ and $\hat{N} \leftarrow K$ if $C = T_0$
        \STATE $j \leftarrow -1$
        \STATE $T_1 = T_0 + \ceil{\hat{N}\log(2T)}$
        \FOR{$t \leftarrow T_0 + 1$ to $T_1$}
            \IF{$j = -1$}
                \STATE Sample $i$ uniformly in $[\hat{N}]$ and pull arm $i$
                \STATE Observe the collision information $\eta(t)$
                \IF{$\eta(t) = 0$}
                    \STATE $j \leftarrow i$
                \ENDIF
            \ELSE
                \STATE Pull arm $j$
            \ENDIF
        \ENDFOR
        \STATE \textbf{Output:} $\hat{N}$, $j$
    \end{algorithmic}
\end{algorithm}

\subsection{Properties}
\label{sect:relaxed-setting-property}
We first show that the strategic player will not affect other players that follow the Musical Chairs approach in \propositionref{prop:musical-chair-robust}.
\begin{proposition} \label{prop:musical-chair-robust}
    If $N - 1$ players follow the Musical Chairs approach, then with probability at least $1 - N / T$, after $T_1 = \ceil{50 K^2\log (4T)} + \ceil{N \log (2T)}$ rounds, regardless of the policy of the strategic player, each player could estimate the number of players $N$ accurately and get a different rank $j \in [N]$ with other players except the strategic player.
\end{proposition}

Similar to \algorithmref{alg:SMAA}, we could show the properties of \algorithmref{alg:SMAA-relaxed} by the following theorem (formal version of \corollaryref{coro:main-relaxed}).
\begin{theorem} \label{thrm:relaxed-setting}
    Under \assumptionref{assum:unique}, suppose $0 < \delta < \delta_0 / 2$. Let $\mu_{\max} = \max_{k \in [K]} \mu_k$. Then \algorithmref{alg:SMAA-relaxed} satisfies the following properties.
    \begin{enumerate}
        \item When all players follow \algorithmref{alg:SMAA-relaxed}, the expected regret for each player $j$ is upper-bounded by the following equation.
            \begin{equation}
                \bbE[\regret_j(T)] \le \sum_{k \not\in \calM^*} \frac{(z^* - \mu_k)(\log T + 4 \log (\log T))}{\kl(\mu_k + \delta, z^* - \delta)} + (50K^2 + N)\mu_{\max}\log (4T) + 10N^3K(13K + \delta^{-2}) + N \mu_{\max}.
            \end{equation}
        \item When all players follow \algorithmref{alg:SMAA-relaxed}, the expected number of rounds that are not a Nash equilibrium demonstrated in \theoremref{thrm:PNE} is upper bounded by the following equation.
            \begin{equation}
                \bbE[\noneq(T)] \le  N\sum_{k \not\in \calM^*} \frac{\log T + 4 \log (\log T)}{\kl(\mu_k + \delta, z^* - \delta)} + (50K^2 + N)\log (4T) + 10N^3K(14K + \delta^{-2}).
            \end{equation}
        \item The policy profile where all players follow \algorithmref{alg:SMAA-relaxed} is an $\epsilon$-Nash equilibrium \textit{w.r.t.} to the game specified by $(\calS_{\policy}, \{\calU_j^{\policy}\}_{j=1}^N)$ and is $(\beta, \epsilon + \beta\gamma)$-stable with
        \begin{equation}
            \begin{aligned}
                \beta = & \, \frac{\delta_0}{z^*}, \\
                \epsilon = & \, \sum_{k \not\in \calM^*} \frac{(z^* - \mu_k)(\log T + 4 \log (\log T))}{\kl(\mu_k + \delta, z^* - \delta)} + (50K^2 + N)\mu_{\max}\log (4T) + 10N^3K(13K + \delta^{-2}) + N \mu_{\max}, \\
                \gamma = & \, \sum_{k \not\in \calM^*}\frac{\log T + \log(\log T)}{\kl(\mu_k + \delta, z^* - \delta)} + (50K^2 + N)\mu_{\max}\log (4T) + 10N^3K(13K + \delta^{-2}) + N \mu_{\max}.
            \end{aligned}
        \end{equation}
        Here $z^*$ is provided in \equationref{eq:z-star}.
    \end{enumerate}
\end{theorem}

\subsection{Proofs}
\subsubsection{Proof of \propositionref{prop:musical-chair-robust}} \label{sect:proof-musical-chair}
\begin{proof}
    At each round $t \in [T_0]$, suppose the probability of the strategic behavior to pull arm $k$ is $q_k(t)$. As a result, the probability of any other player $j$ to observe a collision is
    \begin{equation}
        \begin{aligned}
            P(\eta(t) = 1) & = \sum_{k=1}^K P(\pi_j(t) = k)P(\eta(t) = 1|\pi_j(t) = k) \\
            & = \sum_{k=1}^K \frac{1}{K} \cdot \left(1 - \left(1 - \frac{1}{K}\right)^{N-2}(1 - q_k(t))\right) \\
            & = 1 - \left(1 - \frac{1}{K}\right)^{N-2}\sum_{k=1}^K \frac{1 - q_k(t)}{K} \\
            & = 1 - \left(1 - \frac{1}{K}\right)^{N-1}.
        \end{aligned}
    \end{equation}
    As a result, no matter the behaviors of the strategic player, the probability of other players that follow \algorithmref{alg:musical-chair} to observe a collision is constant. According to Lemma 3 in \citep{rosenski2016multi}, we can get that if $T_0 = \ceil{50\log (4T) K^2}$, then with probability at least $1 - 1 / (2T)$, we have $\hat{N} = N$.
    
    For any $t \in \{T_0 + 1, \dots, T_1\}$, when $\hat{N} = N$, then the probability of a player $j$ that follows \algorithmref{alg:musical-chair} to have $\eta_j(t) = 0$ is at least $1 / N$ since there must exist at least one arm that is not pulled in $[N]$. As a result, when $\hat{N} = N$, since $T_1 - T_0 = \ceil{\hat{N} \log (2T)}$, the probability for player $j$ to always meet collisions is
    \begin{equation}
        \bbP\left[\forall t \in \{T_0 + 1, \dots, T_1\}, \eta_j(t) = 0\right] \le \left(1 - \frac{1}{N}\right)^{T_1 - T_0} \le \exp\left(- \frac{T_1 - T_0}{N}\right) \le \frac{1}{2T}.
    \end{equation}
    As a result, when $T_0 = \ceil{50\log (4T) K^2}$ and $T_1 - T_0 = \ceil{\hat{N} \log (2T)}$, the probability of a player that follows \algorithmref{alg:musical-chair} to estimate $N$ accurately and get a different rank with other players is at least $(1 - 1/(2T))^2 \le 1 - 1/T$. Now the claim follows by applying the union bound for all players.
\end{proof}

\subsubsection{Proof of \theoremref{thrm:relaxed-setting}}

\begin{proof}
    For the first part, let $\calE$ denote the clean event when all players that follow \algorithmref{alg:SMAA-relaxed} estimate $N$ accurately and get a different rank $j$. Let $\mu_{max} = \max_{k \in [K]}\mu_k$ and $\regret_j^{\ee}(T)$ represent the regret in the exploration-exploitation phase (\algorithmref{alg:SMAA}). As a result,
    \begin{equation}
        \begin{aligned}
            & \, \bbE[\regret_j(T)] \\
            = & \, \bbP[\calE]\bbE\left[\regret_j(T) | \calE\right] + \bbP[\bar{\calE}]\bbE\left[\regret_j(T) | \bar{\calE}\right] \\
            \le & \, T_1 \mu_{\max} +  \regret_j^{\ee}(T) + \frac{N}{T} \mu_{\max} \cdot T \\
            \le & \, \sum_{k \not\in \calM^*} \frac{(z^* - \mu_k)(\log T + 4 \log (\log T))}{\kl(\mu_k + \delta, z^* - \delta)} + (50K^2 + N)\mu_{\max}\log (4T) + 10N^3K(13K + \delta^{-2}) + N \mu_{\max}.
        \end{aligned}
    \end{equation}
    
    For the second part, let $\noneq^{\ee}(T)$ represent the rounds of non-equilibrium in the exploration-exploitation phase (\algorithmref{alg:SMAA}). Then we have
    \begin{equation}
        \begin{aligned}
            & \, \noneq(T) \\
            = & \, \bbP[\calE]\bbE\left[\noneq(T) |  \calE\right] + \bbP[\bar{\calE}]\bbE\left[\noneq(T) | \bar{\calE}\right] \\
            \le & \, T_1 + \noneq^{\ee}(T) + \frac{N}{T} \cdot T \\
            \le & \, N\sum_{k \not\in \calM^*} \frac{\log T + 4 \log (\log T)}{\kl(\mu_k + \delta, z^* - \delta)} + (50K^2 + N)\log (4T) + 10N^3K(14K + \delta^{-2}).
        \end{aligned}
    \end{equation}
    
    For the third part, according to \propositionref{prop:musical-chair-robust}, the behaviors of the selfish player in the Musical Chairs phase will not affect the results of other players. In addition, as shown in the proof of \theoremref{thrm:selfish} in \appendixref{sect:proof-selfish}, \algorithmref{alg:SMAA} is not affected by selfish players. As a result, we conclude that the policy profile where all players follow \algorithmref{alg:SMAA-relaxed} is an $\epsilon$-Nash equilibrium with
    \begin{equation}
        \epsilon = \sum_{k \not\in \calM^*} \frac{(z^* - \mu_k)(\log T + 4 \log (\log T))}{\kl(\mu_k + \delta, z^* - \delta)} + (50K^2 + N)\mu_{\max}\log (4T) + 10N^3K(13K + \delta^{-2}) + N \mu_{\max}.
    \end{equation}
    
    Let $\reward^{\ee}_j(T; \bfs)$ represents the reward of player $j$ in the exploration-exploitation phase (\algorithmref{alg:SMAA} part) under the original profile $\bfs$ where all players follow \algorithmref{alg:SMAA-relaxed}. When a selfish player deviate from $s$ to $s'$. Let $\calQ$ be the set of rounds $t$ that player $j$ deviates from the original algorithm in the exploitation-exploration phase and all other players calculate the correct Nash equilibrium (\textit{i.e.}, $\calQ_3$ in the proof of \theoremref{thrm:selfish} in \appendixref{sect:proof-selfish}). The reward of player $j$ and a player $i$ that follows \algorithmref{alg:SMAA-relaxed} is given by the following equations.
    \begin{equation}
        \begin{aligned}
            & \, \bbE[\reward_i(T; s', \bfs_{-j}) - \reward_i(T; \bfs)] \\
            = & \, \bbP[\calE]\bbE[\reward_i(T; s', \bfs_{-j}) - \reward_i(T; \bfs) | \calE] + \bbP[\bar{\calE}]\bbE[\reward_i(T; s', \bfs_{-j}) - \reward_i(T; \bfs) | \bar{\calE}] \\
            \ge & \, -T_1 \mu_{\max} - \bbE[\reward_i^{\ee}(T; s', \bfs_{-j}) - \reward_i^{\ee}(T; \bfs) | \bar{\calE}] - \frac{N}{T} \cdot T \mu_{\max} \\
            \ge & \, -10N^3K(13K + \delta^{-2}) - \sum_{k \not\in \calM^*}\frac{\log T + \log(\log T)}{\kl(\mu_k + \delta, z^* - \delta)} - (50K^2 + N)\mu_{\max}\log (4T) - N \mu_{\max} - z^*\bbE[|\calQ|] \\
            = & \, - \gamma - z^*\bbE[|\calQ|].
        \end{aligned}
    \end{equation}
    And
    \begin{equation}
        \begin{aligned}
            & \, \bbE[\reward_j(T; s', \bfs_{-j}) - \reward_j(T; \bfs)] \\
            = & \, \bbP[\calE]\bbE[\reward_j(T; s', \bfs_{-j}) - \reward_j(T; \bfs) | \calE] + \bbP[\bar{\calE}]\bbE[\reward_j(T; s', \bfs_{-j}) - \reward_j(T; \bfs) | \bar{\calE}] \\
            \le & \, T_1 \mu_{\max} + \bbE[\reward_i^{\ee}(T; s', \bfs_{-j}) - \reward_i^{\ee}(T; \bfs) | \bar{\calE}] + \frac{N}{T}\cdot T\mu_{\max} \\
            \le & \, 10N^3K(13K + \delta^{-2}) + \sum_{k \not\in\calM^*}\frac{(z^*-\mu_k)(\log T + 4\log (\log T))}{\kl(\mu_k+\delta, z^*-\delta)} - \delta_0 \bbE[|\calQ|] + (50K^2 + N)\mu_{\max}\log (4T) + N\mu_{\max} \\
            = & \, \epsilon - \delta_0 \bbE[|\calQ|].
        \end{aligned}
    \end{equation}
    As a result, for any $u \in \bbR^+$, if $\bbE[\reward_i(T; s', \bfs_{-j})] - \bbE[\reward_i(T; \bfs)] \le -u$, we have $\bbE[|\calQ|] \ge (u - \gamma) / z^*$. Then
    \begin{equation}
        \bbE[\reward_j(T; s', \bfs_{-j})] - \bbE[\reward_j(T; \bfs)] \le \epsilon - \delta_0 \frac{u - \gamma}{z^*} = \epsilon + \frac{\delta_0\gamma}{z^*} - \frac{\delta_0}{z^*}u.
    \end{equation}
    Now the claim follows.
\end{proof}

%% file: paragraphs/appendix-auxiliary.tex
\section{Auxiliary Lemmas}
    

\begin{lemma} [Lemma 5 in \citep{combes2015learning}] \label{lemma:number-rounds}
    Let $\calF_t$ be the $\sigma$-algebra generated by $\left\{X_{k}(t), \forall k \in [K]\right\}_{s=1}^t$. Let $j \in [N]$, $k \in [K]$ and $c > 0$. Let $H$ be a random set of rounds such that for all $t$, $\bbI[t \in H] \in \calF_{t-1}$. Assume that there exists $\{C_t\}_{t \in \bbN}$, a sequence of independent binary random variables such that for any $t \ge 1$, $C_t$ is $\calF_t$-measurable and $\bbP[C_t = 1 | t \in H] \ge c$. Further assume for any $t \in H$, $k$ is selected ($\pi_j(t) = k$) if $C_t = 1$. Then
    \begin{equation}
        \sum_{t \ge 1}\bbP[t \in H, \left|\hat{\mu}_{j,k}(t) - \mu_k \right| \ge \delta] \le 2c^{-1}\left(2c^{-1} + \delta^{-2}\right).
    \end{equation}
\end{lemma}

\begin{lemma} [Theorem 10 in \citep{garivier2011kl}] \label{lemma:garivier}
    Let $\{Z_t\}_{t \ge 1}$ be a sequence of independent random variables bounded in $[0, 1]$ with common expectation $\mu = \bbE[Z_t]$. Let $\calF$ be an increasing sequence of $\sigma$-fields such that for each $t$, $\sigma(Z_1, Z_2, \dots, Z_t) \subseteq \calF_t$ and for $s > t$, $Z_s$ is independent from $\calF_t$. Consider a previsible sequence $\{\epsilon_t\}_{t \ge 1}$ of Bernoulli variables (for all $t > 0$, $\epsilon_t$ is $\calF_{t-1}$-measurable). Let $\delta > 0$ and for every $u \in \{1, \dots, t\}$ let
    \begin{equation}
        \begin{aligned}
            S(u) & = \sum_{s=1}^u \epsilon_sX_s, \quad N(u) = \sum_{s=1}^u \epsilon_s, \quad \hat{\mu}(u) = \frac{S(u)}{N(u)}, \\
            b(t) & = \sup \left\{q \ge \hat{\mu}(t): N(t) \kl (\hat{\mu}(t), q) \le \delta \right\}.
        \end{aligned}
    \end{equation}
    Then
    \begin{equation}
        \bbP[b(t) < \mu] \le \lceil \delta \log t \rceil \exp(1-\delta).
    \end{equation}
\end{lemma}